\documentclass[twocolumn]{article}
\usepackage{preprint}
\usepackage{hyperref}
\usepackage[authoryear,round]{natbib}

\hypersetup{
    colorlinks=true,
    linkcolor=blue,
    urlcolor=blue,
    citecolor=blue,
    anchorcolor=black
}
\usepackage[utf8]{inputenc}	
\usepackage[T1]{fontenc}
\usepackage[table,dvipsnames]{xcolor}
\usepackage{lineno}					
\usepackage{tikz} 					
\usetikzlibrary{positioning, arrows.meta, shadows, fit, backgrounds}
\usepackage{fancyhdr}
\usepackage{geometry}
\usepackage{doi}
\usepackage{amssymb}
\usepackage{amsmath}
\usepackage{bbm}
\usepackage{bm}
\usepackage{array,booktabs}
\usepackage{appendix}
\usepackage{authblk}
\usepackage{placeins}
\usepackage{comment}
\usepackage{siunitx}
\usepackage{float}
\usepackage{url}
\usepackage{tabularx}
\usepackage{caption}
\usepackage{subcaption}
\usepackage{makecell}
\usepackage{textcomp}
\usepackage{gensymb}
\usepackage{threeparttable}
\usepackage{cleveref}
\usepackage[euler]{textgreek}
\usepackage{enumitem}
\newcolumntype{Y}{>{\centering\arraybackslash}X}

\usepackage{titlesec}
\titlespacing\section{0pt}{12pt plus 3pt minus 3pt}{1pt plus 1pt minus 1pt}
\titlespacing\subsection{0pt}{10pt plus 3pt minus 3pt}{1pt plus 1pt minus 1pt}
\titlespacing\subsubsection{0pt}{8pt plus 3pt minus 3pt}{1pt plus 1pt minus 1pt}
\makeatletter
\def\@fnsymbol#1{\ensuremath{\ifcase#1\or *\or \dagger\or \ddagger\or
   \mathsection\or \mathparagraph\or \|\or **\or \dagger\dagger
   \or \ddagger\ddagger \else\@ctrerr\fi}}
\makeatother

\newcommand{\mafone}{\ensuremath{\text{MAF}_1}}

\tikzset{
    stackimage/.style={
        inner sep=0pt,
        align=center
    }
}
\tikzset{
    shadow/.style args={#1,#2,#3}{
        opacity=#1,
        xshift=#2,
        yshift=#3
    }
}
\tikzset{
    modelblock/.style={
        draw,                  
        thick,                 
        fill=blue!20,          
        rounded corners=2pt,   
        minimum width=2.8cm,   
        minimum height=2cm,  
        align=center,          
        drop shadow             
    }
}
\tikzset{
    predblock/.style={
        draw,
        rectangle,
        rounded corners=2pt,
        minimum width=2.4cm,
        minimum height=0.6cm,
        align=center,
        fill=orange!20
    }
}
\tikzset{
    opblock/.style={
        draw,
        rectangle,
        rounded corners=2pt,
        fill=yellow!30,
        minimum width=2.5cm,
        minimum height=2cm,
        align=center,
    }
}
\tikzset{
    resblock/.style={
        draw,
        rectangle,
        rounded corners=2pt,
        fill=purple!30,
        minimum width=2.5cm,
        minimum height=2cm,
        align=center,
    }
}

\title{NormalView: tree species classification from backpack and aerial lidar data using geometric projections}

\usepackage{authblk}

\author[1]{Juho Korkeala} 
\author[1\small\ensuremath{\ast}]{Jesse Muhojoki}
\author[1]{Josef Taher}
\author[1]{Klaara Salolahti}
\author[1]{Matti Hyyppä}
\author[1,2]{Antero Kukko}
\author[1,2]{Juha Hyyppä}

\affil[1]{Department of Remote Sensing and Photogrammetry, Finnish Geospatial Research Institute FGI, The National Land Survey of Finland, Vuorimiehentie 5, FI-02150, Espoo, Finland}
\affil[2]{Department of Built Environment, School of Engineering, Aalto University, P.O. Box 11000, FI-00076, Espoo, Finland}

\begin{document}

\twocolumn[\begin{@twocolumnfalse}

\maketitle

\begin{abstract}
Laser scanning has proven to be an invaluable tool in assessing the decomposition of forest environments. Mobile laser scanning (MLS) has shown to be highly promising for extremely accurate, tree level inventory. In this study, we present NormalView, a projection-based deep learning method for classifying tree species from point cloud data. NormalView embeds local geometric information into two-dimensional projections, in the form of normal vector estimates, and uses the projections as inputs to an image classification network, YOLOv11. In addition, we inspected the effect of multispectral radiometric intensity information on classification performance. We trained and tested our model on high-density MLS data (7 species, \(\sim\)\qty{5000}{pts/m^2}), as well as high-density airborne laser scanning (ALS) data (9 species, \(>\)\qty{1000}{pts/m^2}). On the MLS data, NormalView achieves an overall accuracy (macro-average accuracy) of \(95.5\,\%\) \((94.8\,\%)\), and \(91.8\,\%\) \((79.1\,\%)\) on the ALS data. We found that having intensity information from multiple scanners provides benefits in tree species classification, and the best model on the multispectral ALS dataset was a model using intensity information from all three channels of the multispectral ALS. This study demonstrates that projection-based methods, when enhanced with geometric information and coupled with state-of-the-art image classification backbones, can achieve exceptional results. Crucially, these methods rely only on geometric information, and thus are compatible with most sensors. Additionally, we publically release the MLS dataset used in the study, containing 1915 samples.
\end{abstract}


\vspace{0.5cm}

\end{@twocolumnfalse}]

\renewcommand{\thefootnote}{\small\ensuremath{\ast}}
\footnotetext[1]{Corresponding author: Jesse Muhojoki (jesse.muhojoki@nls.fi)}
\renewcommand{\thefootnote}{\arabic{footnote}}
\setcounter{footnote}{0}

\section{Introduction} \label{intro}

Tree species classification is a key tool in understanding forest and urban ecosystems. Understanding the composition of species helps monitor biodiversity, manage forest risks, and to understand the ecological richness of forest environments~\citep{chirici,Gamfeldt2013,HUUSKONEN}. Accurate species classification also has economical impact~\citep{Haara}. In boreal forests, it is found that aspen (\textit{Populus tremula}) holds an important role as a keystone species~\citep{KIVINEN, KOUKI}, serving as a biodiversity hotspot hosting hundreds of species, including woodpeckers and flying squirrels~\citep{Angelstam, Tikkanen, Remm}. In recent years, automated forest inventory has started to replace more traditional methods in forest inventory. Recently, we have seen deep learning-based methods for semantic segmentation of forests (see e.g. \cite{WIELGOSZ2024114367, HENRICH2024102888, RUOPPA2025, xiang2025}). In combination, these segmentation methods and species classification methods can provide us with forest information almost completely automated. 

Mobile laser scanning (MLS) and airborne laser scanning (ALS) have proven to be invaluable tools in forestry. The applications of MLS include estimation of stem curves~\citep{bauwens2016TLS_HH_comp, HyyppaE2020, gollob2020no_gnss}, extracting branch information~\citep{WINBERG2023100040}, and collecting forest inventories~\citep{Liang2014, MOKROS2021}. ALS data has been used for individual tree detection and their characteristics' assessment~\citep{hyyppa1999detecting, Windrim2020_detection}, change detection~\citep{hyyppa2003growth, yu2004automatic, ma2018quantifying, arumae2020thinning, kozniewski2022tracking}, biomass estimations~\citep{Dalponte2018, zhao2018utility} and semantic segmentation of forests~\citep{XIANG2024114078, RUOPPA2025}. Both data types have been used for the classification of tree species~\citep{Puliti,taher2025}. Accurate tree species classification from dense point clouds --- such as MLS or close-range ALS data --- enables collection of reference data efficiently for a large-area forest inventories. The combination of highly accurate stem curves, efficient collection, and tree species classification is required, as the stem curve and taper models are highly dependent on the species \citep{salekin_global_2021, mctague_evolution_2021} and are changing over time due to, e.g.\@ climate change \citep{SCHNEIDER2018}, and the tree stems are not visible from ALS used for large-area laser scanning, and therefore estimating the stem attribute from other attributes (mostly canopy) is required.

To classify trees from point cloud data, various methods have been presented previously. The methods utilising artificial intelligence can be roughly divided into three categories: machine learning based-methods, projection-based deep learning methods, and point-based 3D deep learning methods. Machine learning (ML) methods derive handcrafted features from tree point clouds, such as tree height, crown area, height percentiles  and intensity and reflectance statistics. For an extensive review of various features used in tree species classification, see~\citet{mlfeatures}. Common ML classifiers include random forests~\citep{randomforest}, support vector machines~\citep{cortes1995support}, and linear discriminant analysis~\citep{lda}. These algorithms have been successfully applied in several tree species classification studies, for example by~\citet{HAKULA2023100039,DALPONTE,Orka,TOCKNER,YU2017} for random forests, by~\citet{DALPONTE,Orka} for support vector machines, and by~\citet{Orka,Hardenbol} for linear discriminant analysis.

The recent rapid developments in the field of deep learning (DL) have made DL methods surpass more traditional ML methods, as seen in the benchmarks by~\citet{Puliti} and~\citet{taher2025}. Projection-based methods operate by creating two-dimensional representations of tree point clouds, for example by taking orthographic projection images from multiple angles and colouring the images black-and-white (B\&W)~\citep{Marinelli, Seidel, Blackburn, Puliti, taher2025}, according to depth~\citep{Allen}, intensity~\citep{Blackburn} or geometric features~\citep{Blackburn}. The classification task is largely studied in the field of image-based artificial intelligence, and one benefit of the 2D-models is the performance of state-of-the-art image models, such as YOLO~\citep{YOLO2023}. 

3D-DL models operate directly on the point clouds, taking as input subsampled point coordinates and possible additional attributes, such as intensity and reflectance values. The pioneers of deep learning in the realm of point cloud understanding were PointNet~\citep{qi2017a} and PointNet++~\citep{Qi2017b}. Several studies have shown their ability to classify tree species~\citep{LIU2022a, Liu2022c, Sun, LIU2022b, taher2025, Puliti}. Newer 3D-architectures include PointMLP~\citep{pointmlp}, 3D-CNNs~\citep{pointcnn}, DGCNN~\citep{dgcnn} and PointTransformer~\citep{pointtransformer}. All of these architectures have been utilised to classify tree species: ~\citet{LIU2022b, Liu2022c} used PointMLP, 3D-CNNs were utilised by~\citet{LIU2021, LIU2022b, Puliti}, DGCNN by~\citet{LIU2022b, Puliti}, and PointTransformer in~\citet{Sun, taher2025}. These studies show the prominence of 3D methods in tree species classification.

Recently, two benchmark studies on tree species classification have been published, namely one by~\citet{Puliti} and another by~\citet{taher2025}. The former is carried out on the FOR-species20k\footnote{Available on Zenodo:~\url{https://doi.org/10.5281/zenodo.13255198}} public dataset~\citep{forspecies20k}, and the other on the MS-ALS-SPECIES\footnote{Available on Zenodo:~\url{https://doi.org/10.5281/zenodo.17077256}} dataset~\citep{espooDATA}. The FOR-species20k dataset contains more than \num{20000} samples from 33 species, which are gathered by different methods (terrestrial laser scanning, MLS, unmanned ALS). In their study,~\citet{Puliti} found that image-based models outperformed certain point-based 3D models. In contrast,~\citet{taher2025} found that point-based 3D models, especially transformer-based architectures, outperform image-based models. Hence, it seems currently unclear whether the best classification performance can be achieved by projection-based methods or 3D-architectures.

Normal vector estimations are regularly used in computer vision and point cloud-based applications, for example in segmentation~\citep{YANG}, shape modelling~\citep{Pauly} and point cloud registration~\citep{Serafin}. However, in tree species classification related problems, the use of normal information has not been studied. Normal vectors are used in this study, as they offer essentially the same local geometric information as, say, curvature or other eigenfeatures. Normal vectors are however relatively simple to compute, and descend naturally to RGB images, as explained in Section~\ref{sub:imagecreation}.

To address these gaps, this study investigates the potential of projection-based deep learning methods for tree species classification from point cloud data. We evaluate the suitability of integrating geometric information into projection images through surface normal vector estimation and assess how effectively this geometric cue alone can support species classification compared to radiometry-based approaches. Furthermore, we examine the influence of multispectral intensity information from multiple scanners and test model performance across both mobile and airborne laser scanning datasets.

The original contributions of this study are the following.
\begin{itemize}
    \item We present NormalView, a projection-based deep learning method for classifying tree species from point cloud data. NormalView embeds local geometric information into two-dimensional projections, in the form
    of normal vector estimates, and uses the projections as inputs to an image classification network.
    \item We extensively inspect the effect of radiometric information on tree species classification via projection-based methods. We compare images coloured by intensity to projections done without intensity information, as well as examine the effect of multispectral intensity information compared to single channel intensity information.
    \item We inspect the aptitude of our model on (\(\sim\)\qty{5000}{pts/m^2}) MLS data, as well as high-density ALS data (\(>\)\qty{1000}{pts/m^2}). We compare the performance of our projection-based models on the ALS data with the results of state-of-the-art point-based 3D models \citep{taher2025}, which are trained on the same ALS data.
    \item We release a high-density, close-range, and geometrically highly accurate eight species dataset of MLS tree segments containing 1915 samples \citep{MLSspeciesDATA}. The data is available at~\href{https://doi.org/10.5281/zenodo.21039101}{10.5281/zenodo.21039101}.
\end{itemize}

\section{Materials and Methods} \label{sec:materialsandmethods}

\subsection{Test site} \label{sub:testsite}

\begin{figure*}[!ht]
    \centering
    \includegraphics[width=\textwidth]{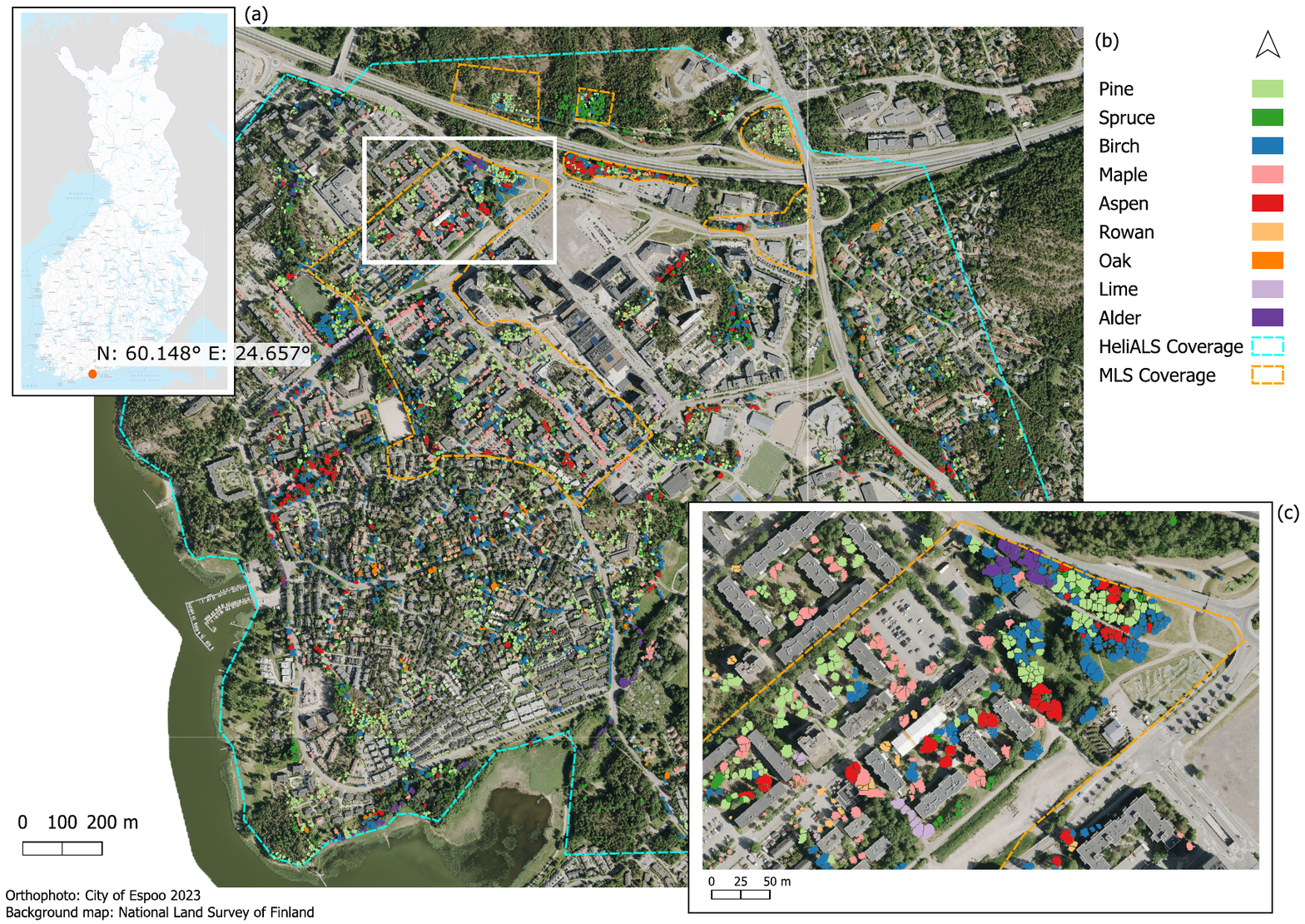}
    \caption{Visualisation of the test site in Espoonlahti. (a): Location of the test site. (b): Orthophoto showing Espoonlahti. The tree segments, as well as the coverage of the different scanning methods, are visible. (c): A zoomed-in view of the orthophoto in (b), showing the tree segments in more detail. Figure adapted from~\citet{taher2025}.}
    \label{fig:test_site}
\end{figure*}

The test site is located in Southern Finland, in Espoonlahti, Espoo (60.148\degree N, 24.657\degree E) and is shown in Figure~\ref{fig:test_site}. The area contains a wide variety of trees in urban, semi-urban, and natural environments. The urban areas contain some planted trees, such as lindens, oaks, and maples. The sites with more naturally occurring trees consist mainly of pines, spruces, birches, and some aspens. The MLS data used in this study contains eight species: pine (\textit{Pinus Sylvestris}), spruce (\textit{Picea sp.}), birch (\textit{Betula sp.}), maple (\textit{Acer platanoides}), aspen (\textit{Populus tremula}), rowan (\textit{Sorbus sp.}),  alder (\textit{Alnus sp.}) and lime (\textit{Tilia sp.}). Alder was not used in the MLS models, as there were too few labelled samples. The ALS data contains nine species: oak (\textit{Quercus robur}) and the eight aforementioned species.

\subsection{Measurements} \label{sub:equipment}

The MLS data was collected using a FARO Orbis laser scanner (FARO Technologies Inc., Lake Mary, Florida, USA). The Faro Orbis is equipped with a Hesai XT32 lidar (Hesai Technology Co., Ltd., Shanghai, China). The measurements were conducted on the 3rd and 11th of June 2024. 19 measurements were made in total, with a single measurement lasting 7--20 minutes. The MLS coverage can be seen in Figure~\ref{fig:test_site}. The laser scanner was mounted on a backpack. The key scanner properties are presented in Table~\ref{tab:scanner_properties}.

The ALS data was gathered in the summer of 2023 by a helicopter equipped with three Riegl (Riegl GmbH, Horn,
Austria) laser scanners: VUX-1HA, miniVUX-1DL, and VQ-840-G. The different scanners operate on different wavelengths, see Table~\ref{tab:scanner_properties}. The flight altitude of the helicopter was approximately \qty{100}{m}. Further information on ALS data collection is available in~\citet{taher2025,espooDATA}.

\begin{table}[ht]
\caption{Scanner Properties.}
\label{tab:scanner_properties}
\scriptsize{
\setlength{\tabcolsep}{3pt}
\begin{threeparttable}
\begin{tabularx}{\columnwidth}{
>{\raggedright\arraybackslash}m{92pt}
*{4}{>{\centering\arraybackslash}Y}
}
    \toprule
     & \multicolumn{3}{c}{ALS} & MLS \\
    \cmidrule(lr){2-4} \cmidrule(lr){5-5}
    Property & VUX-1HA & miniVUX-1DL & VQ-840-G & Hesai XT32 \\
    \midrule
    Divergence $\mathrm{V} \times \mathrm{H}$ (mrad) & 0.5 & $0.5\times1.6$ & 1 & $0.7\times1.7$  \\
    Pulse Repetition Rate (KHz) & 1017 & 100 & 200 & 20 \\
    Range (m at N\% reflectivity) & 85 (10\%) & 140 (20\%) & 150 (20\%) & 80 (10\%) \\
    Wavelength (nm) & 1550 & 905 & 532 & 905 \\
    Beam diameter at ground\tnote{a} (\qty{}{mm}) & 50 & $50\times160$ & 100 & $7\times17$\\
    \bottomrule
\end{tabularx}
\begin{tablenotes}
\item[a] At a \qty{100}{m} (ALS) and \qty{10}{m} (MLS) range from scanners.
\end{tablenotes}
\end{threeparttable}
}
\end{table}

\begin{table*}[ht]
\small
\footnotesize
\caption{Species distribution and model training splits for the different data acquisition methods.}
\label{tab:sp_distribution}
\begin{threeparttable}
\begin{tabularx}{\textwidth}{l *{6}{>{\centering\arraybackslash}X}}
    \toprule
     & \multicolumn{6}{c}{Data type} \\
     \cmidrule(lr){2-7}
     & \multicolumn{3}{c}{ALS} & \multicolumn{3}{c}{MLS} \\
    \cmidrule(lr){2-4}\cmidrule(lr){5-7}
    Species & Train & Test & Total & Train & Test & Total \\
    \midrule
    \multicolumn{7}{l}{\textit{Majority species}} \\
    Pine (Pinus Sylvestris) & 1863 & 465 & 2329 & 639 & 116 & 755  \\
    Spruce (Picea sp.) & 537 & 134 & 671 & 248 & 62 & 310 \\
    Birch (Betula sp.) & 1225 & 307 &1531 & 309 & 65 & 374 \\
    Maple (Acer platanoides) & 543 & 136 & 676 & 188 & 46 &234 \\
    \addlinespace[0.5ex]
    \multicolumn{7}{l}{\textit{Minority species}} \\
    Aspen (Populus tremula) & 377 & 94 & 472 & 87 & 21 & 108 \\
    Rowan (Sorbus sp.) & 163 & 41 & 204 & 48 & 9 & 57 \\
    Oak (Quercus robur) & 53 & 13 & 66 & - & - & - \\
    Lime (Tilia sp.) & 127 & 32 & 158 & 52 & 14 & 66\\
    Alder (Alnus sp.) & 104 & 27 & 131 & - & - & 11\tnote{a} \\
    \midrule
    Total & 4989 & 1249 & 6238 & 1582 & 333 & 1904 \\
    \bottomrule
\end{tabularx}
\begin{tablenotes}
\footnotesize
\item[a] Not used in MLS models.
\end{tablenotes}
\end{threeparttable}
\end{table*}

The field reference was collected using a browser-based solution by the employees of the Finnish Geospatial Research Institute (FGI) in 2024 and 2025. The browser-based application allowed for users to see the segmented trees, overlaid on an orthophoto of the area. With the application, the users could identify the species of the tree on the field, as well as add notes. The application was also used to verify and correct initial labels. The species of some planted trees were identified from the database of~\citet{espoo_open_data}.

\subsection{Data preprocessing} \label{preprocessing}

\begin{figure*}[!ht]
    \centering
    \begin{subfigure}[b]{0.48\textwidth}
        \centering
        \includegraphics[width=\textwidth]{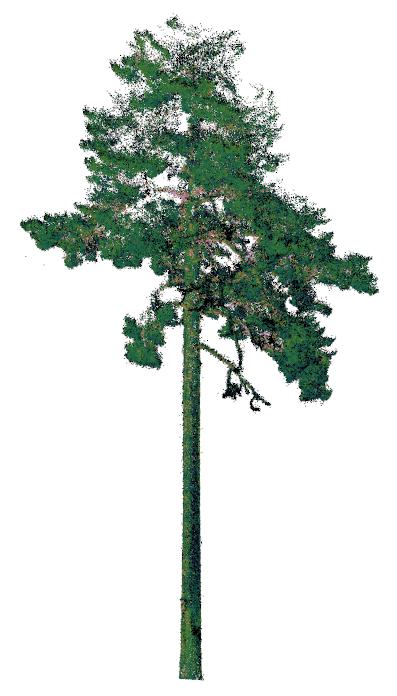}
        \caption{}
        \label{fig:MLStree}
    \end{subfigure}
    \hfill
    \begin{subfigure}[b]{0.48\textwidth}
        \centering
        \includegraphics[width=\textwidth]{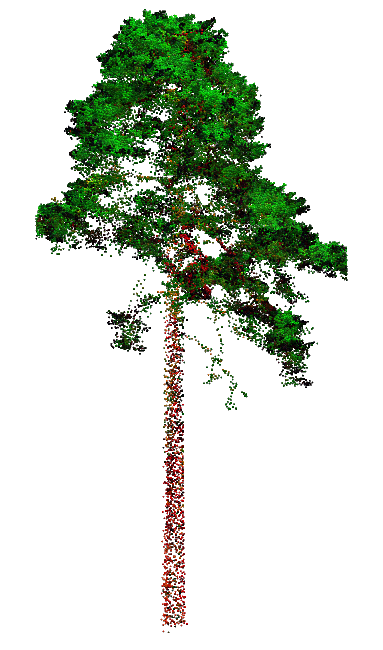}
        \caption{}
        \label{fig:ALStree}
    \end{subfigure}
    \caption{The same pine tree as captured by (a) the backpack mobile scanner, and (b) ALS. The point counts of the segments are \num{399239} and \num{94586} respectively. The MLS point cloud has been coloured by gamma corrected (\(\gamma = 1.15\)) single-channel intensity values (wavelength = \qty{905}{nm}.) The ALS point cloud has been coloured by gamma corrected (\(\gamma = 4\)) intensity values (RGB colour channels: red = \qty{1550}{nm}, green = \qty{905}{nm}, blue = \qty{532}{nm}).}
    \label{fig:MLSALStrees}
\end{figure*}

The preprocessing of the ALS data was conducted on RiProcess (Riegl) and TerraScan (Terrasolid Ltd., Helsinki, Finland) software. Importantly, the data used in this study contains intensity information from each of the three channels for each point. This was achieved by using nearest-neighbour interpolation to fill in missing intensity values. Reflectance was used as the "source" for the intensity in RiProcess, i.e.\@ the intensity values were distance-adjusted. The trees were segmented using a watershed segmentation-based algorithm introduced first in~\citet{yu2010comparison} and tested in~\citet{Kaartinen2012}, and called Local Maxima Finding (FGI\_LOCM). For further information on data processing and reference data collection, we refer the reader to~\citet{taher2025,espooDATA}.

The raw MLS data from the scanner was first processed by the FARO Connect software, which applies Faro's 3D SLAM algorithm. The point clouds were then randomly subsampled to a spatial resolution of \qty{2}{cm}, where the minimum distance between any two points is at least \qty{2}{cm}. Noisy points were removed using a Statistical Outlier Removal -filter (SOR filter). The SOR filter evaluates the average distances to~\textit{k} neighbours for each point and removes points for which the average distance is greater than the average distance of all points plus~\textit{n} times the standard deviation of the average distances. We used \(k = 10\) and \(n = 2\). Then, a Cloth Simulation Filter~\citep{CSF} was applied to the point cloud, to identify the ground points. The measurement sites contained urban elements, such as buildings, cars, and traffic signs, which were manually removed from the point clouds. At this stage, the point clouds contained only the trees. The preprocessing steps were performed using open source CloudCompare software~\citep{CC}.
    
Individual tree segmentation of the MLS trees was done with the TreeIso algorithm~\citep{treeISO}, which is based on a cut-pursuit algorithm for weighted graphs~\citep{landrieu}. The segments produced by TreeIso were manually checked, and unsatisfactory segments were manually corrected using CloudCompare.

To compare the segmented MLS trees with the reference data, they need to be transformed from the local coordinates of the scanner to the global ETRS-TM35FIN coordinate system. To achieve this, we calculated a rigid transformation between the coordinates of each MLS-scan, \(x_\mathrm{local}\), and global ALS coordinates based on anchor points, such as traffic signs, that were found in both point clouds. We calculated an optimal transformation \(x_\mathrm{global} = Rx_\mathrm{local} + t\), where \(R\) is a rotation matrix, \(t\) a translation vector, and \(x_\mathrm{global}\) the new transformed coordinates, that minimised the squared distance between the ALS and MLS anchor points.

After transforming the points to the global coordinates, segmented trees were matched with the reference data. We adopted the same strategy as in~\citet{HAKULA2023100039}, and utilised a Hausdorff distance based closest neighbour algorithm, as presented in~\citet{Yu2006}. Two points (trees), \(a \in A\) and \(b \in B\), in different sets (measurement data and reference data), \(A\) and \(B\), were considered a match if \(a\) is the closest point to \(b\) in \(A\), and \(b\) is the closest point to \(a\) in \(B\). In addition, the distance between the matched points was required to be less than \qty{3}{m} for them to be considered a match. The reference data for the MLS data is the same as for the ALS data.

In total, we arrive at 1915 MLS segments and 6238 ALS segments. The species-wise distributions in the data can be seen in Table~\ref{tab:sp_distribution}. The difference between the data from the different acquisition methods is visualised in Figure~\ref{fig:MLSALStrees}, where the left tree (MLS) has a much higher point density, and especially the trunk and branch structure is clearer. The tree on the right (ALS) on the other hand has comparable, or even more refined, canopy structure, due to it being scanned from above, as opposed to the MLS scans. In the MLS data, there are some trees which were captured by multiple different measurements due to overlapping measurement paths. As distinct scans are inherently different, and hence contain useful unique data, we included all viable scans of such trees in the dataset. There are 178 trees which are present in two segments and 11 trees which are present in three segments. To avoid bias in the model testing, we ensured that all such trees belong to the training set and that the models were not tested on such trees. For each point cloud representing a tree, we evaluated approximate coordinates of the tree trunk. The coordinated were evaluated using a heuristic algorithm, which bins the points in \(x\)- and \(y\)-directions, and calculates a score for each bin based on point density, vertical point spread, vertical gaps, and height. The mean coordinates of the bin with the highest score are the approximate trunk coordinates.

\subsection{Methodology} \label{sub:methodology}

\begin{figure*}[htb]
    \centering
    \begin{subfigure}[b]{0.495\textwidth}
        \centering
        \includegraphics[width=\textwidth]{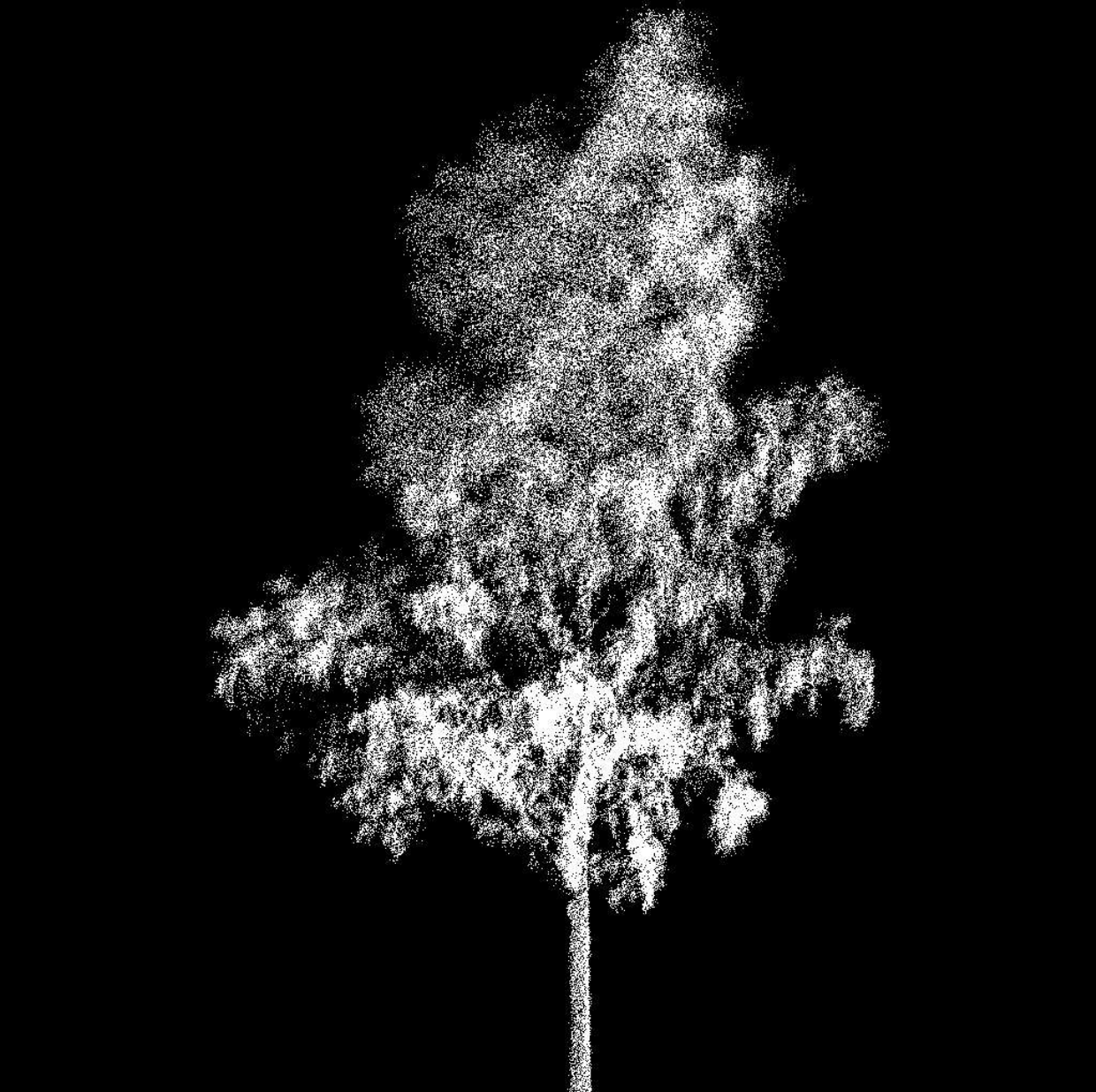}
        \caption{A WOP silhouette image of a birch}
        \label{fig:birch_full}
    \end{subfigure}
    \hfill
    \begin{subfigure}[b]{0.495\textwidth}
        \centering
        \includegraphics[width=\textwidth]{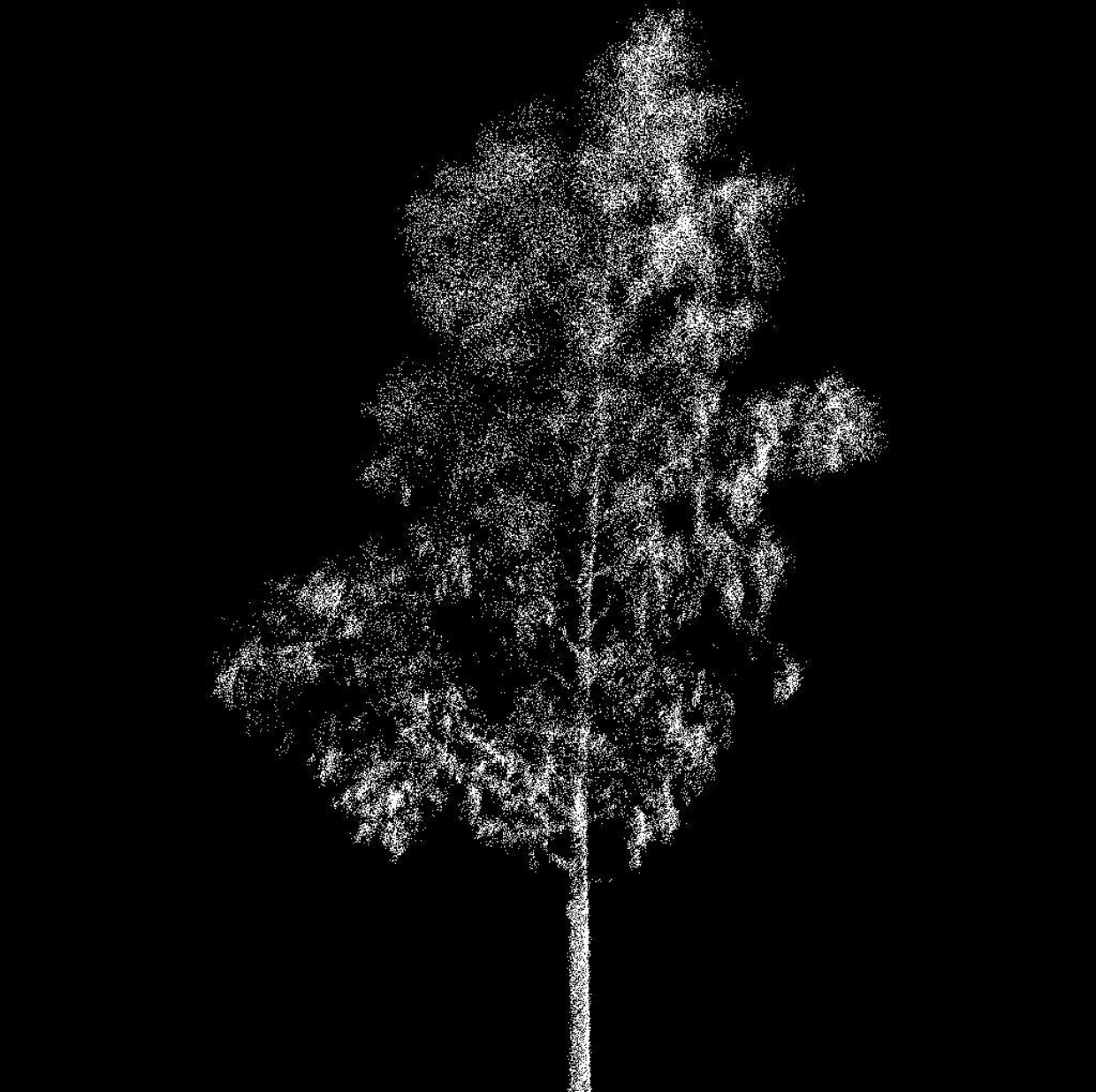}
        \caption{A sliced WOP silhouette image of a birch}
        \label{fig:birch_slice}
    \end{subfigure}
    \par\medskip
    \begin{subfigure}[b]{\textwidth}
        \centering
        \includegraphics[width=\textwidth]{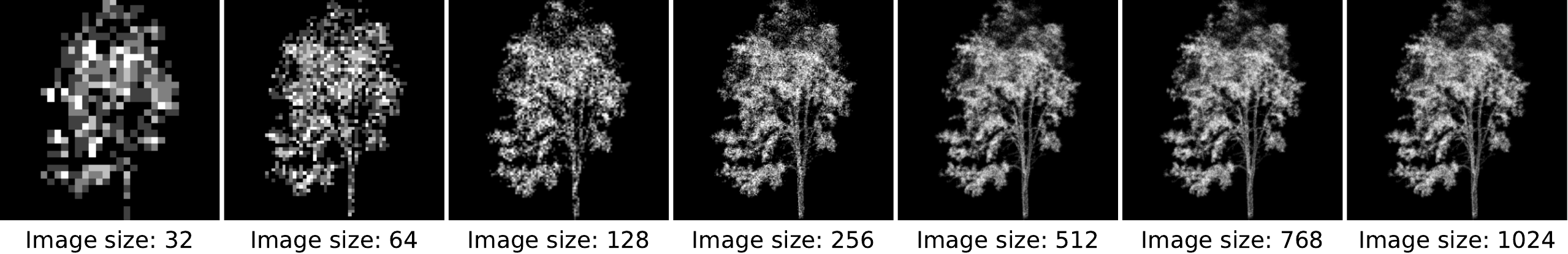}
        \caption{WOP images of the same rowan, scaled down to different sizes.}
        \label{fig:imsizes_rowan}
    \end{subfigure}
    \caption{(a)-(b) Two examples of images used in training the models. The images are WOP images created from the same birch from the same angle. (a) All the points are used for the projection. (b) Points nearest to the "viewing plane" are removed to reduce occlusion on the trunk and inner branches. (c) Images in different resolutions of the same rowan. The resolution is achieved by bilinearly interpolating the image of size \(1024\times1024\).}
    \label{fig:birch_fullvslice}
\end{figure*}

From each tree point cloud, we created projection images by three distinct methods: (1): an orthographic projection with intensity values as colours (OPI), (2): an orthographic projection with RGB-colours coming from normal vector estimations, NormalView (NV), and (3): an orthographic projection where each pixel is either black or white (WOP), which is in essence a silhouette image. Further explanations of the colouring schemes are in Section~\ref{sub:imagecreation}. The models trained on WOP-images serve as a baseline, to see how well the trees can be classified with only projection images, without any radiometric input. With the projected images, we trained a classification model that takes images as input. To this end, we used the YOLOv11x\footnote{\url{https://github.com/ultralytics/ultralytics}} architecture, as provided by Ultralytics (Ultralytics Inc. Frederick, Maryland, USA)~\citep{YOLO2023}.

From the MLS data, we create images according to the three methods mentioned above. From the ALS data we create two image datasets: one with images from seven species to compare the models
trained on ALS data more directly with the models trained on MLS data, and the other with images from the full dataset with nine species, to compare our models with the benchmark in~\citet{taher2025}. Both ALS datasets contain images created by the three methods mentioned above. We analyse the viability of the image creation approaches separately for both the seven and nine species datasets. Finally, we combined the MLS and ALS datasets into a single combined dataset, and created images according to all three methods mentioned above. The purpose of the combined dataset is to analyse the viability of the proposed method on data from varying acquisition sources.

\subsubsection{Image creation} \label{sub:imagecreation}

\begin{figure*}[!ht]
\centering
        \includegraphics[width=\textwidth]{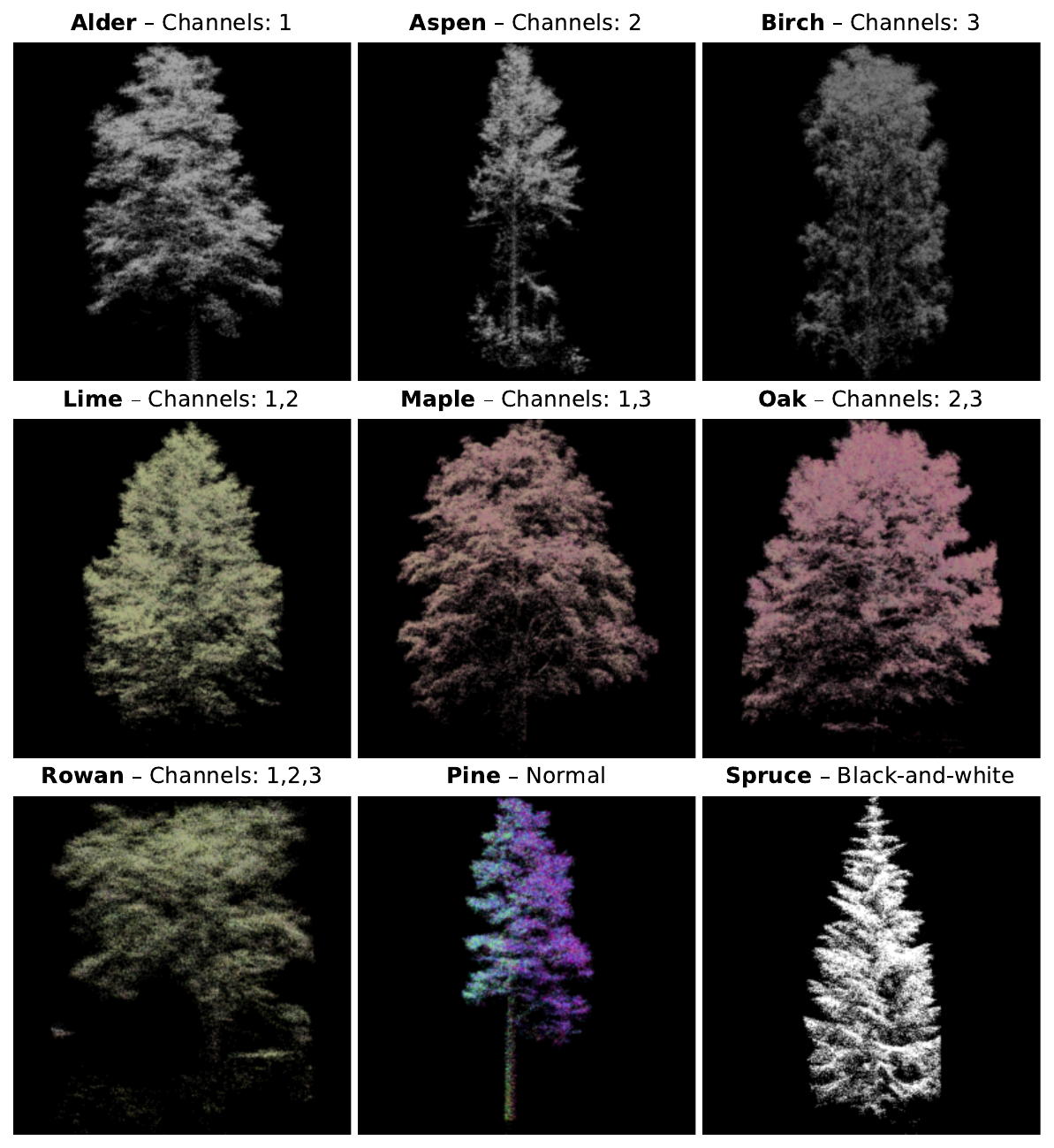}
        \caption{Example images of the nine different tree species and image colouring methods used in model training. All samples are from the ALS data. For explanations for the colours, see section~\ref{sub:imagecreation}.}
        \label{fig:species_images}
\end{figure*}

For all colouring schemes, the orthographic projection images are created by projecting the 3D points of the tree onto a projection plane. The pixel coordinates are obtained by finding the orthogonal, or perpendicular, projection of the point onto the plane. The projection plane is the \(xz\)-plane, and hence the projection corresponds to extracting the \(x\)- and \(z\)-coordinates of the points. The \(x\)- and \(y\)-coordinate axes correspond to easting and northing in the georeferenced data, and the \(z\)-axis to the upwards direction. The projections are taken from multiple angles, as the silhouette of the tree varies depending on the viewing angle. The projections from different angles are achieved by rotating the point cloud around the \(z\)-axis and projecting onto the \(xz\)-plane. Prior to taking the projections, we centered the tree in the origin, by translating the approximated coordinates of the trunk to the origin. In this study, we used five different viewpoints, corresponding to rotations of \(0\degree, 72\degree, 144\degree, 216\degree\) or \(288\degree\) around the \(z\)-axis. In the case that multiple points have the same projected coordinates, and would hence overlap in the projection image, the point nearest to the projection plane was used as the pixel source.

From each angle, we created two images: the full projection containing all points and a sliced image in the depth direction to gain insight of the structure of the tree closer to the trunk. This information includes the growing direction of branches, for example. To generate the sliced images, we excluded all points in the rotated point cloud whose y-coordinate satisfied \(y > t_y + k\), where \(t_y\) is the estimated y-coordinate of the tree trunk, and \(k\) is a used-defined threshold. The remaining points were used to create the image. In this study, we used \(k = \qty{0.7}{m}\). This selection provides useful structural information of the insides of the tree, while being robust against errors in the trunk position estimation. The differences between the full images and the sliced images can be seen in Figures~\ref{fig:birch_full} and~\ref{fig:birch_slice}. Notably, the trunk and inner branches are more visible in the sliced image than in the full image. Consequently, we created ten \(512 \times 512\) images of each tree per colouring scheme, which are discussed below. In addition, to test the effect of image size on classification accuracy, we created images of size \(1024\times1024\) from the MLS data. Multiple images with varying size were produced from the full image via bilinear interpolation, see Figure~\ref{fig:imsizes_rowan}, and performance was tested on each resolution individually.

In the OPI images, we used the intensity values of the points as a single-channel colour input. The intensity values provided by the scanners belong to the interval \([0, \num{65536} ]\) and were normalised by dividing by the maximum value, \num{65536}, since the absolute scale of the intensity can reveal relevant information about the tree. The scaled intensity values were then linearly scaled to the interval \([0, 255]\), and rounded to integers. For the MLS data, these values were then used as input for each of the three 8-bit colour channels in an RGB-image, resulting in a greyscale image. The ALS data contains information from three intensity channels, from which we created intensity-coloured images by using different permutations of the intensities. We created greyscale intensity images from each individual channel as described for the MLS data, and also three types of images containing two different channels, and finally images using all three channels. For the images containing two different channels, the third colour value in the images was set to zero. For the images containing data from all three channels, we performed a similar normalisation as above and used the three channels as input to the red, green, and blue colour channels. The various colouring schemes are illustrated in Figure~\ref{fig:species_images}.

For NV images, we first estimated a normal vector for each point. The normal at a point is estimated by finding the nearest $N$ points of the point in question and fitting a tangent plane via principal component analysis. In this study, we set $N = 20$. This allows for enough points so that the plane is approximately tangential to the surface at the point, while taking points from a small enough area so that we preserve local geometry. In the ablation studies it was found that the choice of \(N=20\) provides the best balance between MLS and ALS data, although the effect of the value of the parameter is rather miniscule, as seen in Table~\ref{tab:ablation_NN}. The unit-length normal vector of this plane was then chosen as the normal. We oriented the normals to point outward, so that the spatial angle between the normal vector and the vector from the point to the estimated trunk is greater than \ang{90}. Subsequently, the components of the normal vector were used as the red, green, and blue colour channels, after a linear scaling to the interval \([0,255]\), and rounding to integer values.

In the WOP images, a pixel was simply coloured white, if there is a projected point corresponding to it. For the intensity and normal images, we applied a Gaussian convolution kernel, to make the images smoother. The convolution kernel is \(3 \times 3\) with a standard deviation of 0.85. No filter was applied to the WOP images.

Projecting the points onto a two-dimensional plane inherently causes information loss, as we lose one degree of spatial freedom. Consequently, we lose intrinsic structural relationships in the data. The projection tactics described in this section attempt to combat this: multiple viewing angles provide a more thorough view of the tree than one singular angle, and the slice images allow the models to "see inside the trees", as can be seen in Figure~\ref{fig:birch_fullvslice}. The key idea behind the normal projections is to embed three-dimensional structural information into the images via colour. Normal vectors describe the local three-dimensional geometry of each point. Another way this could be achieved is by using eigenfeatures derived from the local structural tensor of each point, as is done by~\citet{Blackburn}.

Additionally, we examined the effect of image size in model training on model performance. Image size refers to the parameter passed on to the YOLO model in the training phase. The input images are scaled to the set image size by applying bilinear interpolation. All images were created to have size \(1024\times1024\), and were scaled using the bilinear interpolation to the set size when training the model. Ultralytics implements the scaling using OpenCV's resize-function~\citep{opencv_library}. We tested the effect of image size only on the intensity-coloured images. The differences in perceived visual quality are large for the smaller image sizes, as demonstrated by Figure~\ref{fig:imsizes_rowan}. We see that a significant portion of visual details are lost in images sized \(128\times 128\) and smaller.

\subsubsection{Model training}\label{sub:mod_train}

\begin{table}[ht]

\footnotesize
\centering
\small{
\begin{threeparttable}
\caption{\small{Parameters used in model training. For the entries for image and batch size, the first one refers to ALS models, and the second one to MLS models. The values inside parentheses for RAM and GPU refer to the machine used to train models with image sizes 768 and 1024 in the image size tests.}}
\label{tab:mod_params}
\begin{tabularx}{\columnwidth}{
>{\raggedright\arraybackslash}m{92pt}
*{1}{>{\centering\arraybackslash}Y}
}
    \toprule
    Parameter & Value \\
    \midrule
    Epochs & 400 \\
    Image size & 256 \slash \ 512 \\
    Batch size & 16 \slash \ 5 \\
    Optimizer & AdamW \\
    lr0\tnote{a} & 0.0005 \\
    Dropout & 0.15 \\
    Patience\tnote{b} & 30\\
    Momentum\tnote{c} & 0.8 \\
    Learning rate scheduler & Cosine \\
    Augmentation policy & RandAugment \\
    Python & 3.11. \\
    PyTorch & 2.7.0. \\
    CUDA & 12.4. \\
    RAM & 32 GB (256 GB) \\
    GPU & 12 GB Nvidia RTX A3000 \newline (24 GB Nvidia Quatro RTX A 6000) \\
    
    \bottomrule
\end{tabularx}
\begin{tablenotes}
\footnotesize
\item[a] Initial learning rate.
\item[b] Number of epochs to wait before stopping training early if performance does not improve.
\item[c] $\beta_1$ parameter in AdamW; controls incorporation of past gradients.
\end{tablenotes}
\end{threeparttable}
}
\end{table}

\begin{figure*}[ht]
\centering
\resizebox{\textwidth}{!}{%
\begin{tikzpicture}[
    >=Stealth,
    node distance=2cm,
    img/.style={inner sep=0pt, align=center},
    block/.style={draw, rectangle, minimum width=2.8cm, minimum height=1.2cm, align=center},
    stack/.style={draw, rectangle, minimum width=2.8cm, minimum height=1.2cm},
    label/.style={font=\small}
]
\node[img, font=\Large] (input) {\includegraphics[height=100pt]{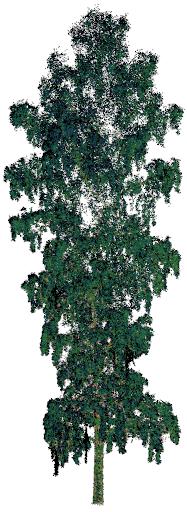}\\Input point cloud};

\node[img, right=of input, font=\Large] (intensity)
{\includegraphics[height=100pt]{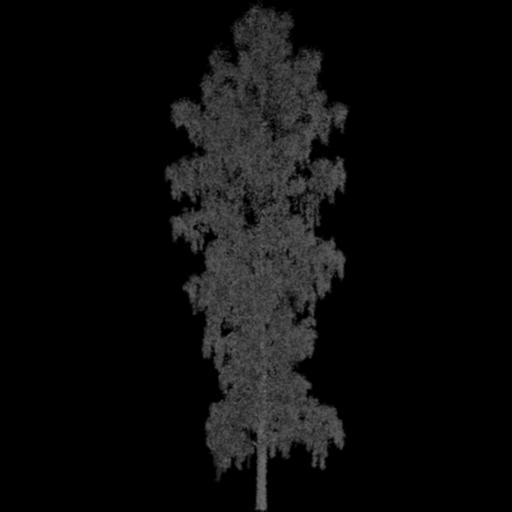}\\OPI};

\node[img, above=0.5cm of intensity, font=\Large] (normals)
{\includegraphics[height=100pt]{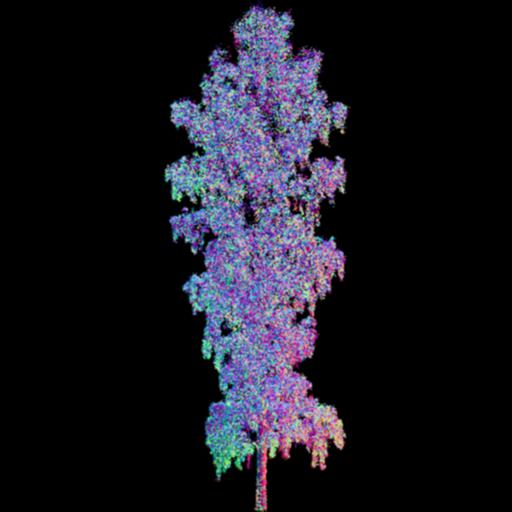}\\NV};

\node[img, below=0.5cm of intensity, font=\Large] (BW)
{\includegraphics[height=100pt]{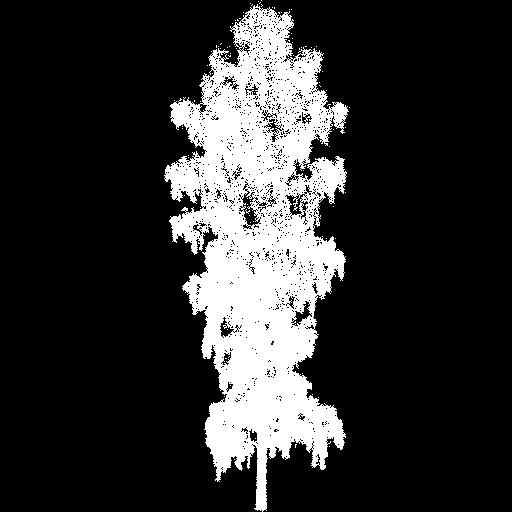}\\WOP};

\draw[->, thick] (input) -- (normals.south west);
\draw[->, thick] (input) -- (intensity.west);
\draw[->, thick] (input) -- (BW.north west);
\begin{pgfonlayer}{background}
\node[draw, rounded corners, fill=green!10, thick, inner sep=5pt, fit=(normals) (intensity) (BW)] (bg_images) {};
\end{pgfonlayer}
\node[above=2mm of bg_images.north, font=\Large\bfseries] {Image Creation};

\node (train_stack_pos) [above right=1.5cm and 4.5 cm of intensity] {};
\node[below=2.6cm of train_stack_pos, font=\Large\bfseries, align=center] {Slice and\\rotate};

\node[stackimage, opacity=0.2, xshift=0.2cm, yshift=0.2cm] (train_shadow1) 
    at (train_stack_pos) {\includegraphics[height=80pt]{figure5b.jpg}};
\node[stackimage, opacity=0.1, xshift=0.4cm, yshift=0.4cm]  (train_shadow2)
    at (train_stack_pos) {\includegraphics[height=80pt]{figure5b.jpg}};
\node[stackimage] (train_main) at (train_stack_pos) 
{\includegraphics[height=80pt]{figure5b.jpg}};

\draw[->, thick] (normals.east) -- (train_main.west);
\draw[->, thick] (BW.east) -- (train_main.west);
\draw[->, thick] (intensity.east) -- (train_main.west);

\node (aug_stack_pos) [right=4cm of train_main] {};

\node[stackimage, opacity=0.3, xshift=0.2cm, yshift=0.2cm] (train_shadow_aug1) 
    at (aug_stack_pos) {\includegraphics[height=80pt]{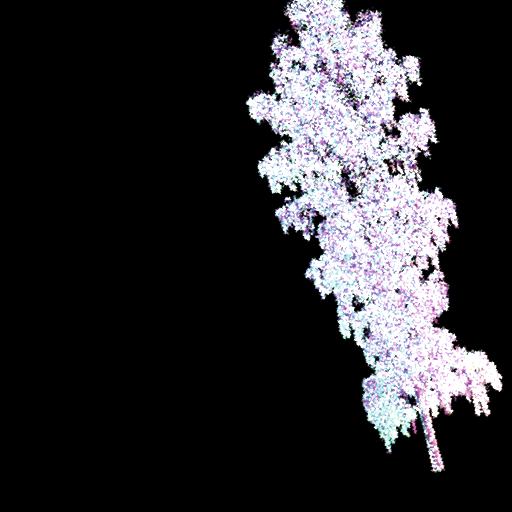}};
\node[stackimage, opacity=0.2, xshift=0.4cm, yshift=0.4cm]  (train_shadow_aug2)
    at (aug_stack_pos) {\includegraphics[height=80pt]{figure5e.jpg}};
\draw[->, gray, thick] (train_shadow1) -- (train_shadow_aug1.west);
\draw[->, gray, thick] (train_shadow2) -- (train_shadow_aug2.west);

\node[stackimage] (train_main_aug) at (aug_stack_pos) {\includegraphics[height=80pt]{figure5e.jpg}};

\node[above=0.1cm of train_shadow2, align=center, font=\Large] (train_label) {10 images};
\node[above=0.1cm of train_shadow_aug2, align=center, font=\Large] (aug_label) {Image augmentations};

\node[modelblock, right=2.5cm of train_main_aug, font=\Large] (train_model)
{YOLOv11x};
\node[right=1.25cm of aug_label, align=center, font=\Large] (tmodel_label) {Train model};

\draw[->, thick] (train_main) -- (train_main_aug.west);

\draw[->, thick] (train_main_aug) -- (train_model.west);
\draw[->, gray, thick] (train_shadow_aug1) -- (train_model.west);
\draw[->, gray, thick] (train_shadow_aug2) -- (train_model.west);

\begin{pgfonlayer}{background}
\node[draw, rounded corners, fill=red!10, thick, inner sep=5pt, fit=(train_main) (train_label) (train_main_aug) (train_model)] (bg_train) {};
\end{pgfonlayer}
\node[above=2mm of bg_train.north, font=\Large\bfseries] {Training};

\node (infer_stack_pos) [below right=2cm and 4.5cm of intensity] {};

\node[stackimage, opacity=0.3, xshift=0.2cm, yshift=0.2cm] (infer_shadow1) 
    at (infer_stack_pos) {\includegraphics[height=80pt]{figure5b.jpg}};
\node[stackimage, opacity=0.2, xshift=0.4cm, yshift=0.4cm]  (infer_shadow2)
    at (infer_stack_pos) {\includegraphics[height=80pt]{figure5b.jpg}};

\node[stackimage] (infer_main) at (infer_stack_pos) 
{\includegraphics[height=80pt]{figure5b.jpg}};

\node[above=0.1cm of infer_shadow2, align=center, font=\Large] (infer_label) {50 images};
    
\node[modelblock, right=3cm of infer_stack_pos, font=\Large] (infer_model)
{YOLOv11x};
\node[right=1.25cm of infer_label, align=center, font=\Large, yshift=-0.1cm] (imodel_label) {Predict species\\from each image};

\draw[->, thick] (normals.east) -- (infer_main.west);
\draw[->, thick] (intensity.east) -- (infer_main.west);
\draw[->, thick] (BW.east) -- (infer_main.west);
\draw[->, thick] (infer_main) -- (infer_model.west);
\draw[->, gray, thick] (infer_shadow1) -- (infer_model.west);
\draw[->, gray, thick] (infer_shadow2) -- (infer_model.west);

\node[predblock, above right=1cm and 2cm of infer_model] (pred1) {Probabilities 1};
\node[predblock, below=0.1cm of pred1] (pred2) {Probabilities 2};
\node[predblock, below=0.1cm of pred2] (pred3) {Probabilities 3};

\node[align=center, scale=1.5] (pred_dots) [below=0.0cm of pred3] {$\displaystyle \vdots$};

\node[predblock] (pred_bottom) [below=0.15cm of pred_dots] {Probabilities 50};

\node[opblock, below right=1cm and 2.cm of pred1, font=\Large] (aggregate)
{Aggregated\\species\\probabilities};

\node[resblock, right=2.5cm of aggregate, font=\Large] (result)
{Prediction};

\node[above = 0.9cm of aggregate, align=center, font=\Large] (agg_label) {Sum probabilities};
\node[right=1.55cm of agg_label, align=center, yshift=-0.3cm, font=\Large] (pred_label) {Choose species\\with highest\\probability};

\draw[->, thick] (infer_model.east) -- (pred1.west);
\draw[->, thick] (infer_model.east) -- (pred2.west);
\draw[->, thick] (infer_model.east) -- (pred3.west);
\draw[->, thick] (infer_model.east) -- (pred_bottom.west);
\draw[->, thick] (pred1.east) -- (aggregate.west);
\draw[->, thick] (pred2.east) -- (aggregate.west);
\draw[->, thick] (pred3.east) -- (aggregate.west);
\draw[->, thick] (pred_bottom.east) -- (aggregate.west);
\draw[->, thick] (aggregate) -- (result) node[midway, above] {$\displaystyle \operatorname*{arg\,max}_s \, (p_s)$};

\begin{pgfonlayer}{background}
\node[draw, rounded corners, fill=brown!10, thick, inner sep=5pt, fit=(infer_main) (pred1) (pred_bottom) (result) (infer_label) (pred_label)]
(bg_infer) {};
\end{pgfonlayer}
\node[above=2mm of bg_infer.north, font=\Large\bfseries] {Inference};

\end{tikzpicture}
}
\caption{A flowchart representing the workflow of the used methods. First, we create images from the input point cloud (green). Then, we either train a model using the created images (red), or use a trained model to predict the species of the input (brown).}
\label{fig:flowchart}

\end{figure*}

As the classification model, we used the YOLOv11x architecture, provided by Ultralytics~\citep{YOLO2023}. The model is pretrained on the ImageNet dataset~\citep{ImageNet}. The training parameters can be seen from Table \ref{tab:mod_params}. We divided the data so that approximately 80\,\% of the data is used for model training, and 20\,\% is used for testing. \citet{taher2025} use a 5-way k-fold split in their studies. The train/test split we used corresponds to the train/test split of the first fold in~\citet{taher2025}.

In the training phase, we used early stopping to prevent overfitting and to save time. Early stopping means stopping training prior to reaching the set number of epochs, if no improvement in validation metrics is observed for a certain number of epochs. Ultralytics provides this functionality via the \texttt{patience} parameter, which was set to 30. Inference is done by creating images from 25 different angles, and aggregating the results to arrive at the final prediction, which is the species with the highest aggregated probability. For each angle, we create the full and sliced images, so 50 images in total are used to predict the species. This method of inference adds a significant degree of robustness to the prediction, at the cost of computational time.

Image augmentations in the training phase of image classification models are recognised to be of significant benefit to model robustness and to avoid overfitting~\citep{Shorten}. The augmentations can include geometric transformations (translations, scaling, and horizontal flips), colour transformations (hue, saturation, and brightness adjustments), blocking parts of the image, and combining multiple images into one. We used RandAugment~\citep{randaugment}, as the image augmentation policy. The implementation is embedded into YOLOv11 by Ultralytics, and their implementation relies on the implementation by PyTorch\footnote{\url{https://docs.pytorch.org/vision/main/generated/torchvision.transforms.RandAugment.html}.Visited on 27.08.2025.}. The used software versions and hardware specs can be seen in Table~\ref{tab:mod_params}. In the image size tests, models using sizes 768 and 1024 were trained on a more powerful machine, due to high VRAM consumption.

The workflow regarding the methods outlined above is shown in Figure~\ref{fig:flowchart}. From the input data, we create images according to the methods described in Subsection~\ref{sub:imagecreation}. From these images, we can either train a new model, or predict the species of the input point cloud using a trained model. Notably, in the training phase we create 10 images per tree point cloud, but when doing inference, 50 images are created.

\subsection{Experiments} \label{sub:experiments}

Below are listed the experiments that were conducted in this study. 

\begin{itemize}[left=2pt]
\setlength{\itemsep}{0pt}
\item Comparison of models trained on OPI, NV, and WOP images created from MLS data.
\item Comparison of models trained on OPI, NV and WOP images created from ALS data using seven species and the full nine species data set.
\item Comparison of models trained on OPI, NV and WOP images created from a mixed dataset consisting of both the MLS and ALS datasets.
\item Conducting ablation studies on the effects of point density, nearest neighbour parameter in normal vector estimation, and slice images on model perforamance.
\item Evaluating the effect of image size on model performance.
\end{itemize}

To compare the models using different image types, we trained models with the same parameters and evaluated their performance using the metrics listed in section~\ref{sub:mod_eval}. The OPI images created from ALS data contain all seven permutations of different intensity channels as described above. The motivation behind these tests is to evaluate the performance of NormalView against models using radiometric information, as well as to inspect the differences between models using intensity information from a single source compared to models using data from multiple sources.

Additionally, we examined the effect of image size in model training on model performance. Image size refers to the parameter passed on to the YOLO model in the training phase. The input images are scaled to the set image size by applying bilinear interpolation. All images were created to have size \(1024\times1024\), and were scaled using the bilinear interpolation to the set size when training the model. Ultralytics implements the scaling using OpenCV's resize-function~\citep{opencv_library}. We tested the effect of image size only on the intensity-coloured images. The differences in perceived visual quality are large for the smaller image sizes, as demonstrated by Figure~\ref{fig:imsizes_rowan}. We see that a significant portion of visual details are lost in images sized \(128\times 128\) and smaller. However, the visual differences are not as prevalent between the larger images. The aim was to find out how much the image size parameter affects model performance, and what is the optimal image size.

When training on the mixed MLS + ALS dataset, we used image size 256 in training. As discussed above, the images are scaled to the set resolution before being passed on to the model. Thus, nothing is done to the ALS images, as they already have resolution \(256 \times 256\), but the MLS images are scaled down from their \(512 \times 512\) resolution. Further, on the mixed dataset we used the data from scanner 2 (miniVUX-1DL) as the intensity data for the ALS segments. The reason for this choice is that out of the three models using single-channel intensity data on the nine species ALS data, the one using channel 2 data performed the best, see Table~\ref{tab:als79_mod_res}.

\subsection{Model evaluation}\label{sub:mod_eval}

To evaluate the trained models, we utilised four macro-level metrics: overall accuracy (OA), macro-average accuracy (MAA), macro-average \(\text{F}_1\)-score (MA\(\text{F}_1\)), Cohen's kappa (\textKappa)~\citep{kappa}, and three species-level metrics: precision (P), recall (R) and species-wise \(\text{F}_1\) (\(\text{F}_1\)). We denote true positives \(\text{TP}_s\), true negatives \(\text{TN}_s\), false negatives \(\text{FN}_s\), and false positives \(\text{FP}_s\), where \(s\) denotes the species.

Overall accuracy is defined as the ratio of correct predictions over all test samples

\begin{equation*}
    \text{OA} = \frac{1}{N_{test}}\sum_{i=1}^{N_ {test}}\mathbbm{1}(\hat{y}_i = y_i),
\end{equation*}

where $N_{test}$ is the number of samples in the test set, $\mathbbm{1}$ is the indicator function, $\hat{y}_i$ is the predicted label of the $i$th segment, and $y_i$ is the correct label of the $i$th segment. OA is a straightforward metric for classification models, but can be a poor indicator of performance, for example if the underlying class distribution is imbalanced~\citep{HE}.

The species-wise precision, recall, and $\text{F}_{1}$-score are defined as

\begin{equation*}
    \mathrm{P}_s = \frac{\mathrm{TP}_s}{\mathrm{TP}_s + \mathrm{FP}_s}, \quad
    \mathrm{R}_s = \frac{\mathrm{TP}_s}{\mathrm{TP}_s + \mathrm{FN}_s}, \quad
    \mathrm{F}_{1,s} = \frac{2\mathrm{P}_s\mathrm{R}_s}{\mathrm{P}_s + \mathrm{R}_s}
\end{equation*}

To gain insight of model performance in the presence of the aforementioned imbalance in the species distribution, we also calculated the macro-average accuracy for each model. MAA is more sensitive to model behaviour on the minority species than OA, as each species has equal weight. In addition, we evaluated the macro-average $\text{F}_1$-score. The MAA is defined as the arithmetic mean of species-wise recalls, while the MA$\text{F}_1$ is defined as the arithmetic mean of species-wise $\text{F}_1$ scores.

\begin{equation*}
    \mathrm{MAA} = \frac{1}{|S|} \sum_{s \in S} \mathrm{R}_s, \quad
    \mathrm{MAF}_1 = \frac{1}{|S|} \sum_{s \in S} \mathrm{F}_{1,s}
\end{equation*}

 where $S$ denotes the set of all species in the dataset and $|S|$ the number of species in the set. The Kappa score is defined as 
 \begin{equation*}
     \text{\textKappa} = \frac{p_o - p_e}{1 - p_e},
 \end{equation*}
 where $p_o$ is the empirical agreement observed between the predictions and the ground truth and $p_e$ is the proportion of samples agreed upon if the predictions were given by chance.

\section{Results and Discussion} \label{sec:resultsanddiscussion}

In this section, we evaluate the performance of models and provide an analysis of the results. In Section \ref{sub:Model_perf}, we show the performance of the models trained on MLS data and seven species, and on ALS data, both with seven and nine species. Section~\ref{sub:imsize} presents our analysis of the effect of image size on model performance.

\subsection{Model performance evaluation} \label{sub:Model_perf}

\begin{figure*}[!ht]
\captionsetup[subfigure]{skip=1pt, singlelinecheck=false}

    \centering
    \begin{subfigure}[b]{\textwidth}
        \centering
        \caption{}
        \includegraphics[width=\textwidth]{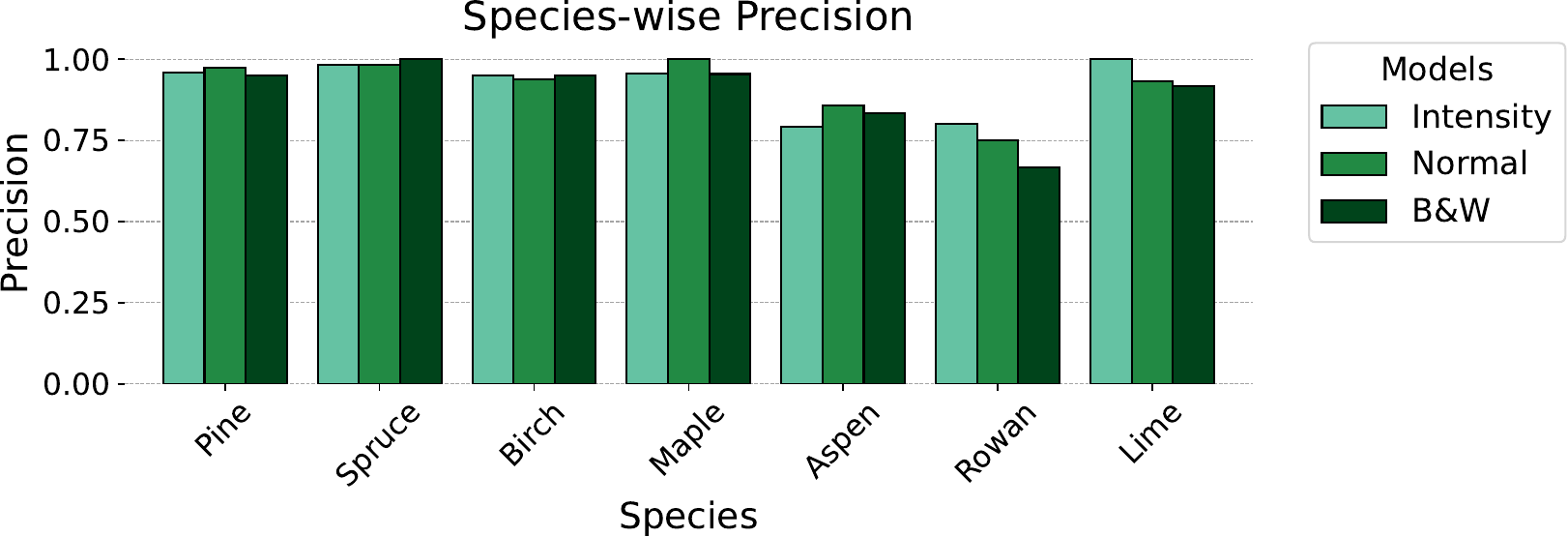}
        \label{fig:MLSprecbar}
    \end{subfigure}

    \vspace{-20pt} 
    
    \begin{subfigure}[b]{\textwidth}
        \centering
        \caption{}
        \includegraphics[width=\textwidth]{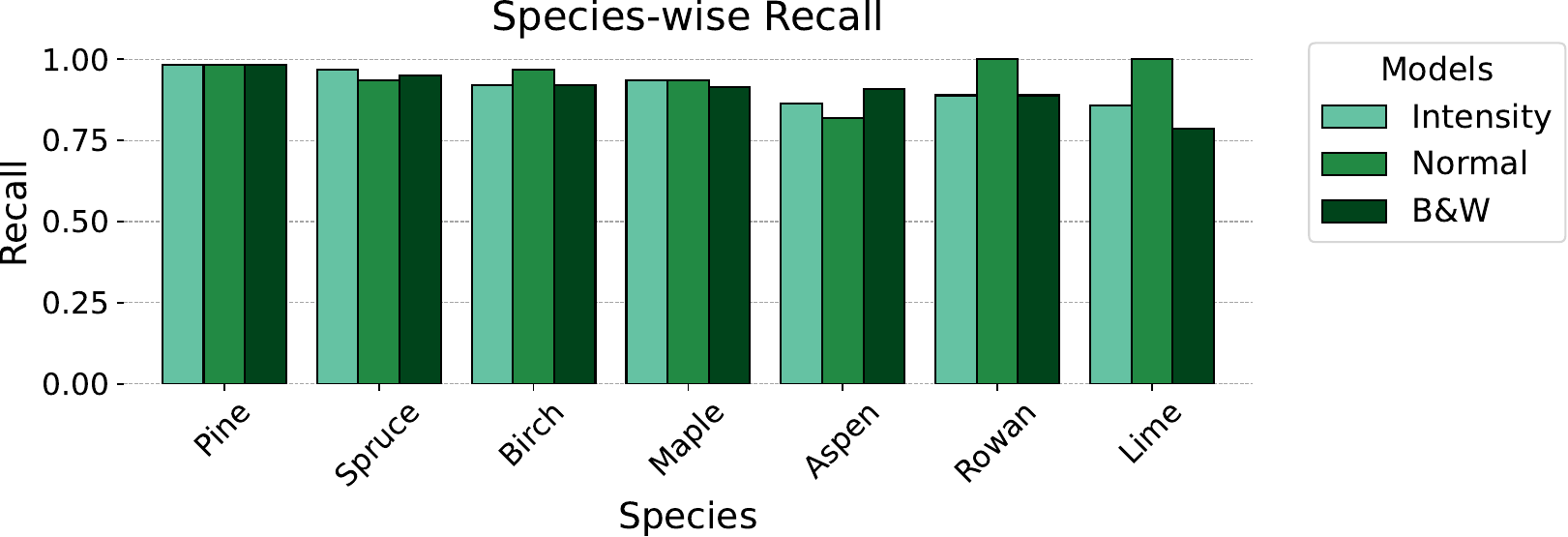}
        \label{fig:MLSrecbar}
    \end{subfigure}

    \vspace{-20pt}
    
    \begin{subfigure}[b]{\textwidth}
        \centering
        \caption{}
        \includegraphics[width=\textwidth]{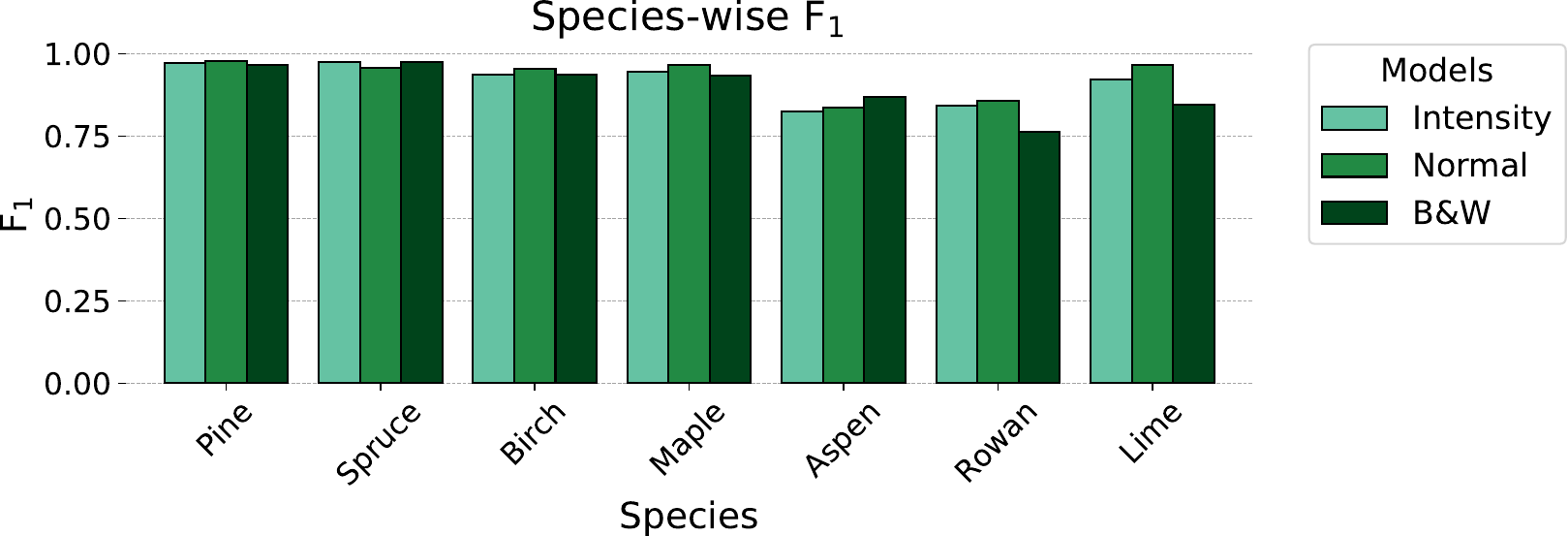}
        \label{fig:f1MLS}
    \end{subfigure}
    \caption{Species-wise metrics of the models trained on MLS data. (a) Species-wise precision, (b) species-wise recall, (c) species-wise \(\text{F}_1\) scores. B\&W refers to the model trained on black-and-white images.}
    \label{fig:MLS_species_metrics}
\end{figure*}

\begin{figure*}[p]
\captionsetup[subfigure]{skip=1pt, singlelinecheck=false}

    \centering
    \begin{subfigure}[b]{0.387\textwidth}
        \centering
        \caption{}
        \includegraphics[width=\textwidth]{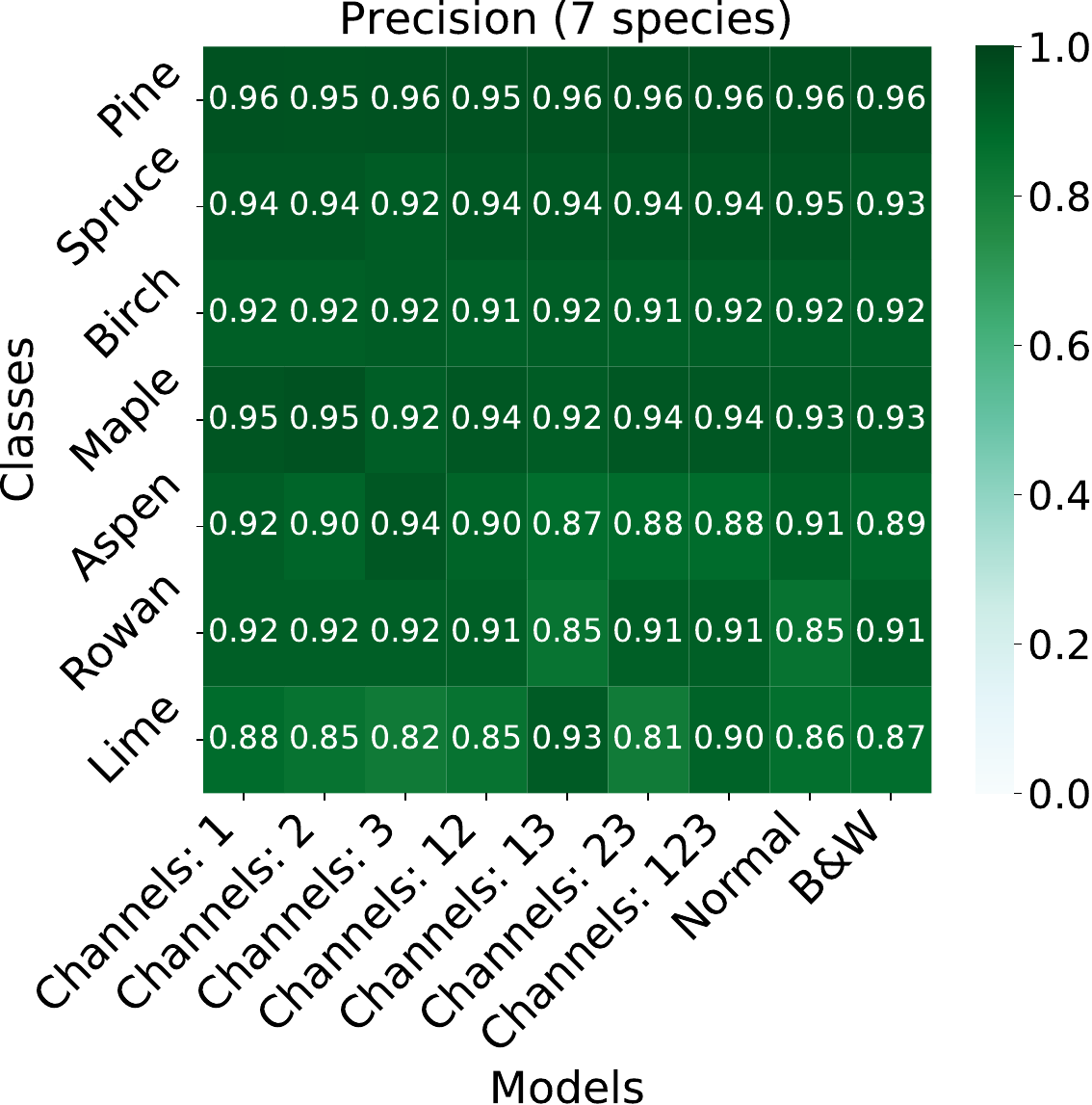}
        \label{fig:ALSprec}
    \end{subfigure}
    \hspace{20pt}
    \begin{subfigure}[b]{0.387\textwidth}
        \centering
        \caption{}
        \includegraphics[width=\textwidth]{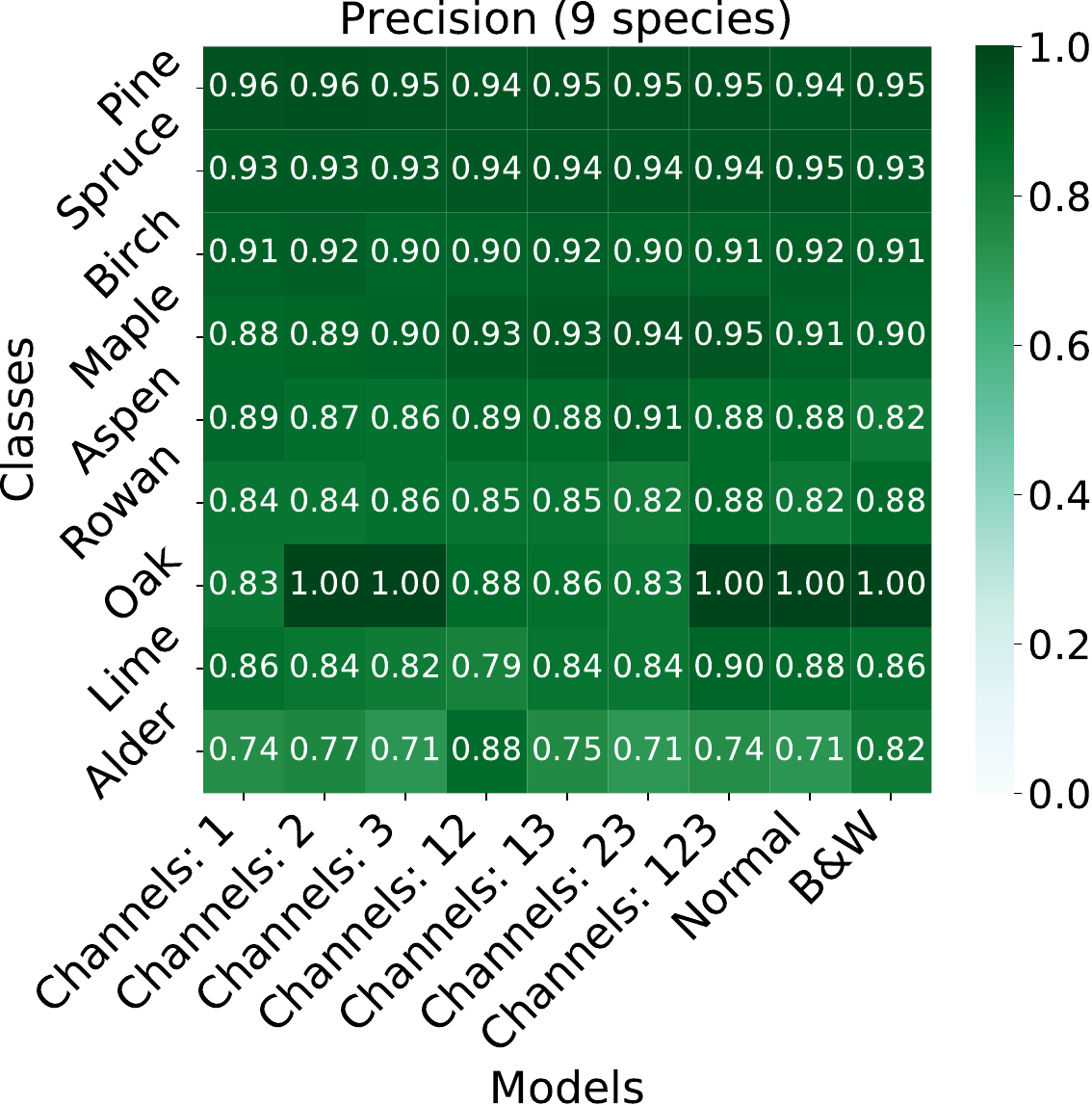}
        \label{fig:ALSprec9}
    \end{subfigure}

    \vspace{-16pt}
    
    \begin{subfigure}[b]{0.387\textwidth}
        \centering
        \caption{}
        \includegraphics[width=\textwidth]{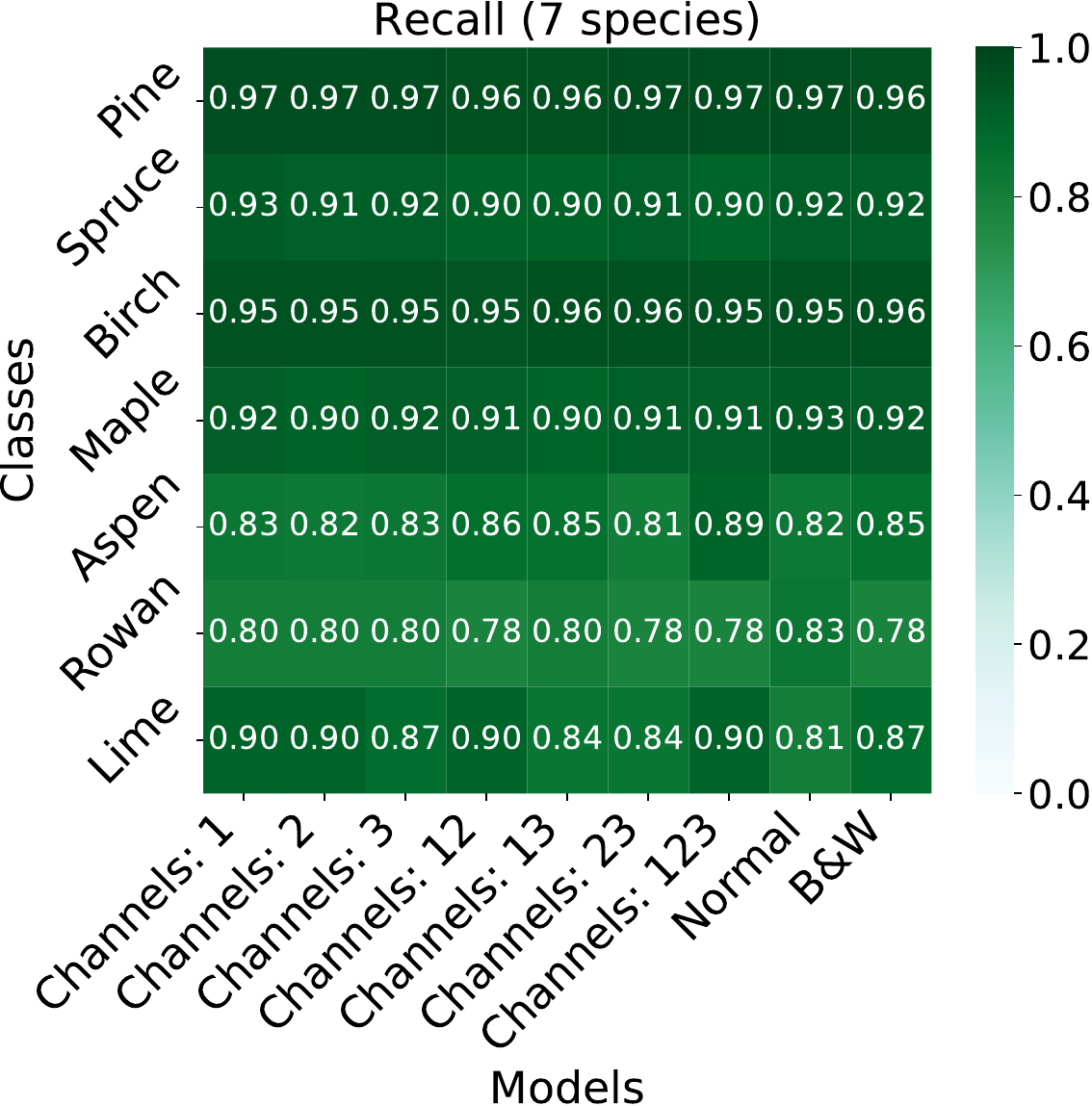}
        \label{fig:ALSrec}
    \end{subfigure}
    \hspace{20pt}
    \begin{subfigure}[b]{0.387\textwidth}
        \centering
        \caption{}
        \includegraphics[width=\textwidth]{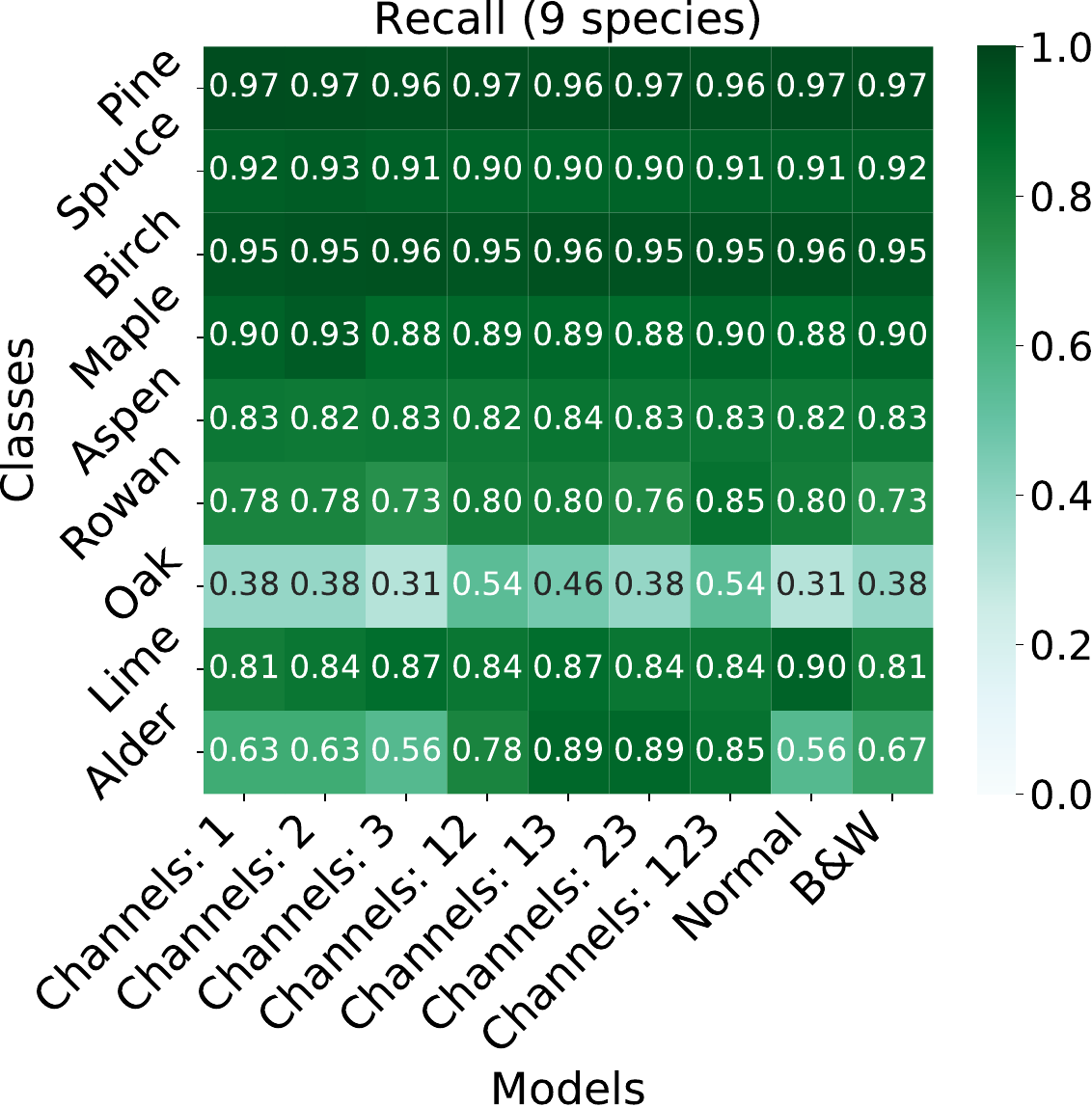}
        \label{fig:ALSrec9}
    \end{subfigure}

    \vspace{-16pt}
    
    \begin{subfigure}[b]{0.387\textwidth}
        \centering
        \caption{}
        \includegraphics[width=\textwidth]{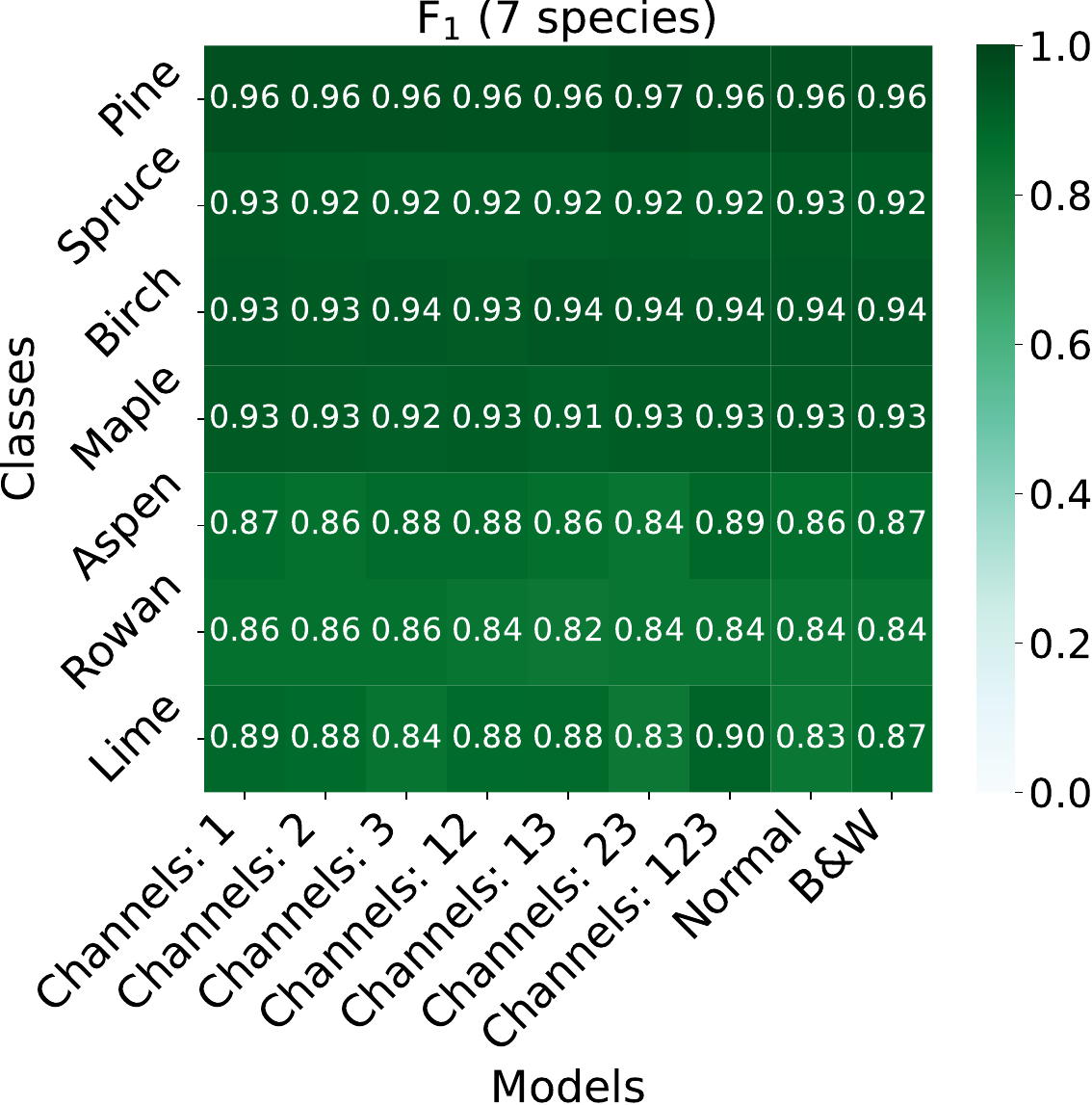}
        \label{fig:f1ALS}
    \end{subfigure}
    \hspace{20pt}
    \begin{subfigure}[b]{0.387\textwidth}
        \centering
        \caption{}
        \includegraphics[width=\textwidth]{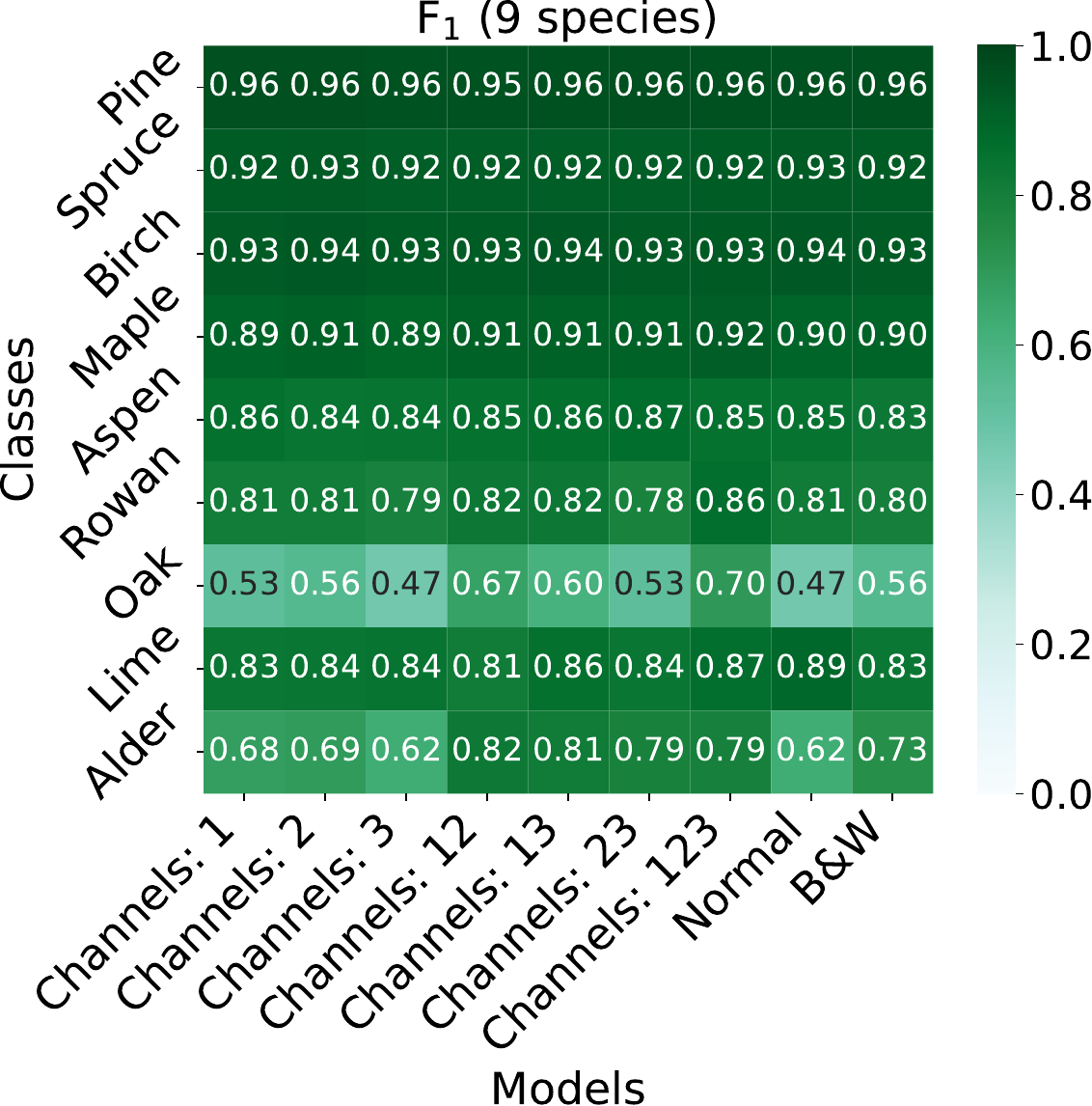}
        \label{fig:f1ALS9}
    \end{subfigure}
    \caption{Species-wise precision, recall, and \(\text{F}_1\)-scores for models trained on ALS data with 7 (left) and 9 (right) species: (a)–(b) precision, (c)–(d) recall, (e)–(f) \(\text{F}_1\)-scores.}
    \label{fig:ALS_species_metrics}
\end{figure*}

The macro-level model performance metrics for the models trained on MLS data are presented in Table~\ref{tab:MLS_mod_res}, and the results for the ALS models are visible in Table~\ref{tab:als79_mod_res}, for both seven and nine species. The species-level metrics are visible in Figures~\ref{fig:MLS_species_metrics} and~\ref{fig:ALS_species_metrics}, for the MLS and ALS models respectively. In the following,  Channel 1 refers to intensity information
provided by VUX-1HA, Channel 2 to miniVUX-1DL, and Channel 3 to VQ-840-G.

\begin{table*}[!t]
\caption{Performance metrics of models trained on MLS data. Bolded values indicate the best performance in each metric.}
\label{tab:MLS_mod_res}
\begin{tabularx}{\textwidth}{>{\raggedright\arraybackslash}X Y Y Y Y}
    \toprule
    Type & OA\,(\%) & MAA\,(\%) & \mafone\,(\%) & \textKappa\,(\%) \\
    \midrule
    Intensity & 94.6 & 91.6 & 91.7 & 93.1\\
    Normal & \textbf{95.5} & \textbf{94.8} & \textbf{93.1} & \textbf{94.2} \\
    Black-and-white & 94.0 & 90.7 & 89.8 & 92.3 \\ 
    \bottomrule
\end{tabularx}
\end{table*}
\begin{table*}[!t]
\caption{Comparison of model performance metrics on ALS data for 7 and 9 species. 
Bolded values indicate the best performance in each metric respectively for 7 and 9 species. 
Channel 1 refers to intensity information provided by VUX-1HA, 
Channel 2 to miniVUX-1DL, and Channel 3 to VQ-840-G.}
\label{tab:als79_mod_res}
\renewcommand{\arraystretch}{1.1}
\begin{tabularx}{\textwidth}{
  >{\raggedright\arraybackslash}m{84pt}
  *{8}{>{\centering\arraybackslash}Y}
}
\toprule
& \multicolumn{4}{c}{\textbf{7 Species}} & \multicolumn{4}{c}{\textbf{9 Species}} \\
\cmidrule(lr){2-5} \cmidrule(lr){6-9}

\multicolumn{1}{>{\raggedright\arraybackslash}X}{Type}
 & \multicolumn{1}{>{\centering\arraybackslash}Y}{\footnotesize OA\,(\%)}
 & \multicolumn{1}{>{\centering\arraybackslash}Y}{\footnotesize MAA\,(\%)}
 & \multicolumn{1}{>{\centering\arraybackslash}Y}{\footnotesize \mafone\,(\%)}
 & \multicolumn{1}{>{\centering\arraybackslash}Y}{\footnotesize \textKappa\,(\%)}
 & \multicolumn{1}{>{\centering\arraybackslash}Y}{\footnotesize OA\,(\%)}
 & \multicolumn{1}{>{\centering\arraybackslash}Y}{\footnotesize MAA\,(\%)}
 & \multicolumn{1}{>{\centering\arraybackslash}Y}{\footnotesize \mafone\,(\%)}
 & \multicolumn{1}{>{\centering\arraybackslash}Y}{\footnotesize \textKappa\,(\%)} \\

\midrule
Channels: 1   & \textbf{93.6} & 90.1 & \textbf{91.1} & \textbf{91.5} & 91.8 & 79.7 & 82.4 & 89.2 \\
Channels: 2   & 93.2 & 89.5 & 90.5 & 91.0 & 92.2 & 80.3 & 83.1 & 89.8 \\
Channels: 3   & 93.4 & 89.5 & 90.3 & 91.2 & 91.2 & 77.9 & 80.8 & 88.5 \\
Channels: 12  & 93.1 & 89.6 & 90.5 & 90.9 & 91.8 & 83.2 & 85.4 & 89.3 \\
Channels: 13  & 93.1 & 88.8 & 89.9 & 90.8 & 92.3 & 84.2 & 85.3 & 90.0 \\
Channels: 23  & 93.2 & 88.3 & 89.4 & 91.0 & 92.1 & 82.3 & 83.6 & 89.6 \\
Channels: 123 & \textbf{93.6} & \textbf{90.6} & \textbf{91.1} & \textbf{91.5} & 
                 \textbf{92.5} & \textbf{84.9} & \textbf{86.8} & \textbf{90.2} \\
Normal        & 93.4 & 88.9 & 89.9 & 91.2 & 91.8 & 79.1 & 81.8 & 89.2 \\
Black-and-white & 93.5 & 89.5 & 90.5 & 91.3 & 91.6 & 79.5 & 82.9 & 89.0 \\
\bottomrule
\end{tabularx}
\end{table*}

\begin{table*}[!t]
    \caption{Performance metrics of models trained on MLS and ALS data. Bolded values indicate the best performance in each metric.}
\label{tab:MLSALS_mod_res}
\begin{tabularx}{\textwidth}{>{\raggedright\arraybackslash}X Y Y Y Y}
    \toprule
    Type & OA\,(\%) & MAA\,(\%) & \mafone\,(\%) & \textKappa\,(\%) \\
    \midrule
    Intensity & 92.2 & \textbf{78.1} & \textbf{81.3} & 89.7\\
    Normal & 92.1 & 76.4 & 79.5 & 89.6 \\
    Black-and-white & \textbf{92.4} & 77.2 & 81.0 & \textbf{90.0}   \\ 
    \bottomrule
\end{tabularx}
\end{table*}


\begin{figure*}[!ht]
\captionsetup[subfigure]{skip=1pt, singlelinecheck=false}

    \centering
    \begin{subfigure}[b]{\textwidth}
        \centering
        \caption{}
        \includegraphics[width=\textwidth]{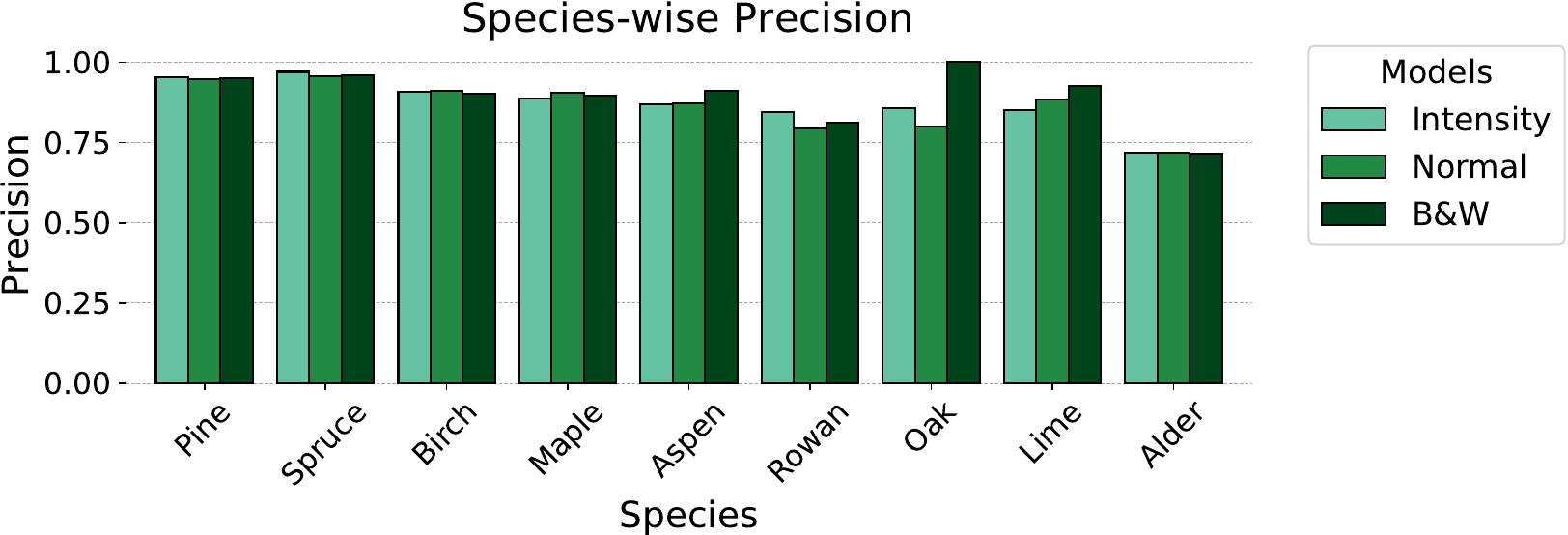}
        \label{fig:MLSALSprecbar}
    \end{subfigure}

    \vspace{-20pt} 
    
    \begin{subfigure}[b]{\textwidth}
        \centering
        \caption{}
        \includegraphics[width=\textwidth]{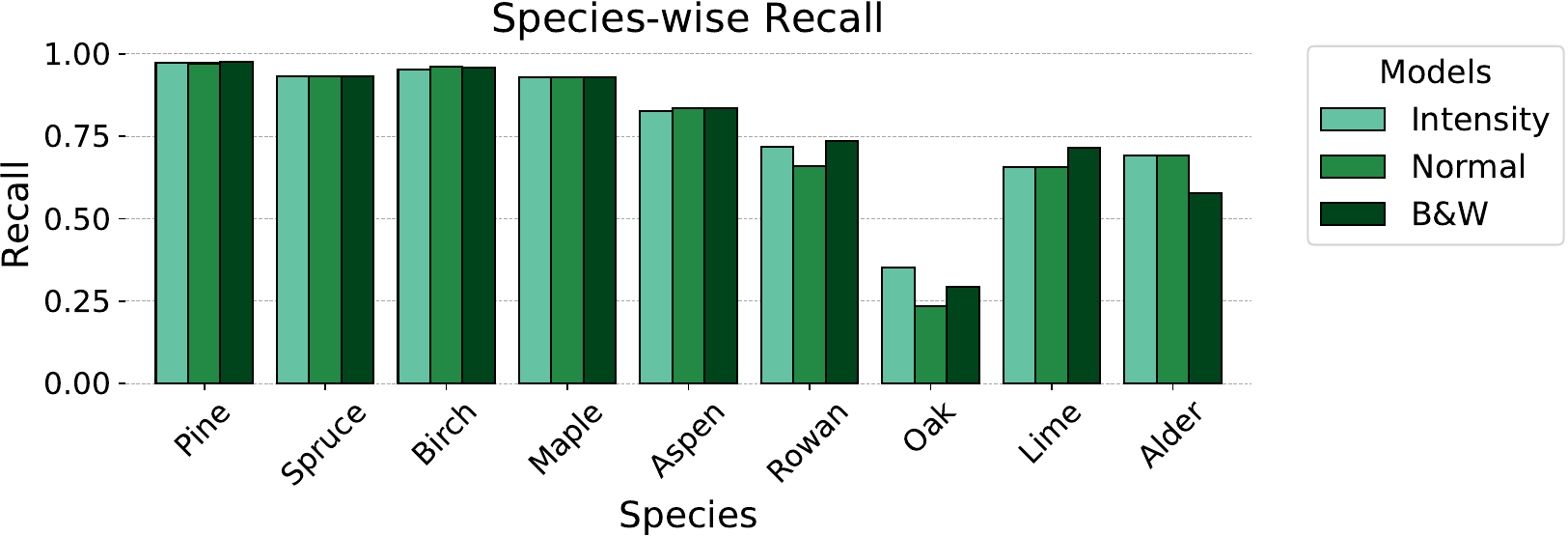}
        \label{fig:MLSALSrecbar}
    \end{subfigure}

    \vspace{-20pt}
    
    \begin{subfigure}[b]{\textwidth}
        \centering
        \caption{}
        \includegraphics[width=\textwidth]{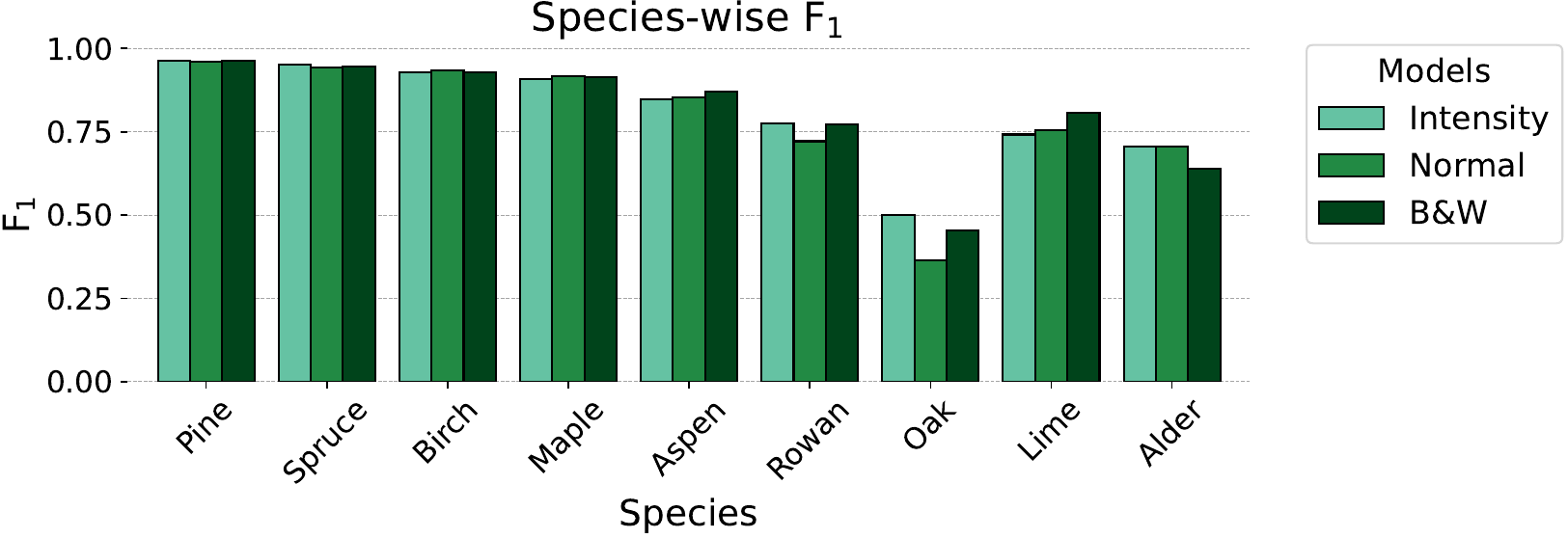}
        \label{fig:f1MLSALS}
    \end{subfigure}
    \caption{Species-wise metrics of the models trained on the mixed MLS + ALS data. (a) Species-wise precision, (b) species-wise recall, (c) species-wise \(\text{F}_1\) scores. B\&W refers to the model trained on black-and-white images.}
    \label{fig:MLSALS_species_metrics}
\end{figure*}

NormalView demonstrates the strongest performance on the MLS data, as can be seen in Table~\ref{tab:MLS_mod_res}. It achieves the best macro-level performance metrics across the board. The strong results in MAA and MA\(\text{F}_1\) highlight the ability of NormalView to take into account even minority species in dense MLS data. This observation is supported by the species-wise performance presented in Figure~\ref{fig:MLS_species_metrics}. NormalView achieves the highest \(\text{F}_1\) for all species except for spruce and aspen.

The intensity-based model performs better than the black-and-white model, but the differences in performance are marginal. For the majority species, the species-wise performance is similar across the models. However, as Figure~\ref{fig:MLS_species_metrics} shows, the differences start to show for the minority species. The black-and-white model performs notably worse than the other two models on rowan and lime.

The overall metrics for species classification for the ALS data are slightly lower than on the MLS data, as illustrated in Table~\ref{tab:als79_mod_res}. This can likely be attributed to the higher point density and superior segment quality in the MLS data. The results on the seven species ALS data remain strong, however, with the OA (MAA) being 92.6\,\% (88.3\,\%) or higher for each model. The average \(\text{F}_1\)-scores and Cohen's kappas all also exceed 90\,\%, except for the macro-\(\text{F}_1\) scores for the normal model, and the models using intensity channels 1 and 3, and 2 and 3. On all the models, the MAA is noticeably lower than OA.

On the nine species dataset, the OAs (MAAs) are 91.6\,\% (77.9\,\%) or higher for each model. There is more notable separation between the OA and MAA for the models trained on nine species, when compared to the MLS models or models trained on seven species. On nine species, the models using multiple intensity channels perform more consistently than the others.

Overall on the ALS data, the best performing model is the one using all three intensity channels, achieving an OA (MAA) of 93.6\,\% (90.6\,\%) on seven species, and 92.5\,\% (84.9\,\%) on nine species. On seven species it is followed closely by the model using only intensity channel 1, which achieved similar performance in all other macro-level metrics, except having an MAA of 90.1\,\%, which is the second highest of the models. On nine species, the second best performer is the model using intensity channels 1 and 3. The overall accuracies of all models are within a percentage point for seven species, and within 1.3 percentage points for nine species. There is a bit more separation in the MAA and MA\(\text{F}_1\) metrics, which imply that there are more noticeable differences in how different models are able to account for minority species.

The difficulty of predicting minority species can be seen in Figure~\ref{fig:ALS_species_metrics}: the metrics are lower for aspen, rowan, and lime, and especially for oak and alder. On the majority species, especially on pine and birch, the species-wise metrics are extremely strong. The model using all intensity channels has the most balanced performance across species. The models using two intensity channels perform extremely similarly on both the seven species and nine species datasets, and on the nine species dataset they outperform the single-channel and geometry-based models. This, and especially the adeptness of the three-channel model, indicate that multispectral information is beneficial on datasets with numerous species.

With the seven species dataset, NormalView and the black-and-white model perform comparatively with all the other models, with the MAAs losing out by less than two percentage points compared to the best model. On nine species, they are notably outperformed by the models using multiple intensity channels. Their performance is on an equal footing with the models using a single intensity channel. Their macro-level metrics are substantially worse than those of the three-channel model. However, the OAs (MAAs) of NormalView and the black-and-white model, \(91.8\,\%\) (\(79.1\,\%\)) and \(91.6\,\%\) (\(79.5\,\%\)) respectively, indicate strong performance, only slightly losing out to the best model (FGI-PointTransformer-DL-3D) in~\citet{taher2025}, which is trained on the same data, and uses full radiometric information.

Table~\ref{tab:MLSALS_mod_res} presents the results of the models trained on the combined MLS and ALS dataset. The differences between the three projection methods are very small. All models achieve an overall accuracy of approximately 92~\%, with differences below 0.3 percentage points. Similarly, the macro-average metrics differ by less than two percentage points. The results are very similar to the results on the full ALS data in Table~\ref{tab:als79_mod_res}. These results suggest that the variability introduced by combining data from two fundamentally different acquisition modalities has a greater influence on classification performance than the choice of image representation. Classfication of the minority species, especially oak, and to a lesser degree rowan, lime, and alder, is again difficult, as suggested by Figure~\ref{fig:MLSALS_species_metrics}.

Interestingly, the black-and-white model performs on par, if not better, with the intensity and NormalView models on the mixed data. This suggests that silhouette information captures structural characteristics that generalise well across both acquisition modalities. The silhouette information remains more consistent between the data acquisition methods, than the local geometry used by NormalView, or the intensity information. In contrast, intensity values and local surface geometry appear to be more sensor-dependent, making their advantages less pronounced when MLS and ALS observations are combined into a single dataset.

\begin{table*}[!t]
\caption{Performance metrics of NormalView trained with varying \(N\) in normal vector estimation.}
\label{tab:ablation_NN}
\renewcommand{\arraystretch}{1.1}
\begin{tabularx}{\textwidth}{
  >{\raggedright\arraybackslash}m{40pt}
  *{8}{>{\centering\arraybackslash}Y}
}
\toprule
& \multicolumn{4}{c}{\textbf{MLS}} & \multicolumn{4}{c}{\textbf{ALS}} \\
\cmidrule(lr){2-5} \cmidrule(lr){6-9}

\multicolumn{1}{>{\raggedright\arraybackslash}X}{NN}
 & \multicolumn{1}{>{\centering\arraybackslash}Y}{\footnotesize OA\,(\%)}
 & \multicolumn{1}{>{\centering\arraybackslash}Y}{\footnotesize MAA\,(\%)}
 & \multicolumn{1}{>{\centering\arraybackslash}Y}{\footnotesize \mafone\,(\%)}
 & \multicolumn{1}{>{\centering\arraybackslash}Y}{\footnotesize \textKappa\,(\%)}
 & \multicolumn{1}{>{\centering\arraybackslash}Y}{\footnotesize OA\,(\%)}
 & \multicolumn{1}{>{\centering\arraybackslash}Y}{\footnotesize MAA\,(\%)}
 & \multicolumn{1}{>{\centering\arraybackslash}Y}{\footnotesize \mafone\,(\%)}
 & \multicolumn{1}{>{\centering\arraybackslash}Y}{\footnotesize \textKappa\,(\%)} \\

\midrule
 3   & 95.5 & 95.1 & 94.6 & 94.2 & 91.2 & 78.0 & 80.9 & 88.5 \\
 5   & 94.6 & 92.7 & 92.9 & 93.0 & 91.8 & 79.5 & 81.9 & 89.2 \\
 10   & 94.3 & 93.6 & 92.3 & 92.7 & 91.7 & 78.9 & 81.5 & 89.1 \\
 15  & 94.3 & 92.9 & 91.7 & 92.7 & 91.1 & 77.8 & 80.3 & 88.4 \\
 20  & 95.5 & 94.8 & 93.1 & 94.2 & 91.8 & 79.1 & 81.8 & 89.2  \\
 30  & 92.8 & 91.2 & 89.7 & 90.7 & 91.8 & 78.8 & 80.4 & 89.2 \\
 40 & 94.5 & 92.1 & 91.6 & 92.7 & 91.7 & 78.7 & 80.4 & 89.1 \\

\bottomrule
\end{tabularx}
\end{table*}
\begin{table*}[!t]
\caption{Metrics of NormalView trained on MLS data with varying point densities. Values in parentheses indicate the difference relative to the NormalView model trained on the full-density MLS dataset (Table~\ref{tab:MLS_mod_res}).}
\label{tab:point_density}
\begin{tabularx}{\textwidth}{
    >{\raggedright\arraybackslash}p{4.2cm}
    Y Y Y Y}
    \toprule
    Point Density (\qty{}{pts/m^2}) & OA\,(\%) & MAA\,(\%) & \mafone\,(\%) & \textKappa\,(\%) \\
    \midrule
    5000 & 94.6 ($-0.9$) & 92.2 ($-2.6$) & 91.3 ($-1.8$) & 93.1 ($-1.1$)\\
    4000 & 93.4 ($-2.1$) & 90.4 ($-4.4$) & 89.1 ($-4.0$) & 91.5 ($-2.7$)\\
    3000 & 93.7 ($-1.8$) & 90.1 ($-4.7$) & 89.6 ($-3.5$) & 91.9 ($-2.3$)\\
    2000 & 94.9 ($-0.6$) & 91.7 ($-3.1$) & 92.1 ($-1.0$) & 93.4 ($-0.8$)\\
    1000 & 94.0 ($-1.5$) & 90.0 ($-4.8$) & 90.5 ($-2.5$) & 92.3 ($-1.9$)\\
    500 & 93.1 ($-2.4$) & 89.9 ($-4.9$) & 89.4 ($-3.7$) & 91.1 ($-3.1$)\\

    \bottomrule
\end{tabularx}
\end{table*}

\subsection{Ablation studies}\label{subsec:ablation_studies}

\begin{table*}[!t]
\caption{Performance of models trained on MLS data without slice images.  Values in parentheses indicate the difference relative to the models trained on the full-density MLS dataset (Table~\ref{tab:MLS_mod_res}).}
\label{tab:no_slice}
\begin{tabularx}{\textwidth}{>{\raggedright\arraybackslash}X Y Y Y Y}
    \toprule
    Type & OA\,(\%) & MAA\,(\%) & \mafone\,(\%) & \textKappa\,(\%) \\
    \midrule
    Intensity & 92.3 ($-2.3$) & 90.6 ($-1.0$) & 90.7 ($-1.0$) & 90.7 ($-2.3$)\\
    Normal & 93.7 ($-1.8$) & 91.1 ($-3.7$) & 90.2 ($-2.9$) & 91.9 ($-2.3$) \\
    Black-and-white & 94.0 ($+0.0$) & 92.9 ($+2.2$) & 91.5 ($+1.7$) & 92.3 ($+0.0$) \\ 
    \bottomrule
\end{tabularx}
\end{table*}

We conducted ablation studies to inspect the effect of different parameters and data quality on the performance of the models. Namely, we tested how the nearest-neighbour parameter \(N\) in normal vector estimation affects the models, do the sliced images provide benefits as opposed to using no sliced images, and how significantly model performance drops, if at all, on lower-density MLS data.

As discussed in Section~\ref{sub:imagecreation}, to estimate the normal vector at a given point, we fit a plane to the nearest \(N\) points, and take the normal vector of the plane to be the normal vector at the point in question. To investigate the effect of this parameter, we created images by using a varying \(N\) in the normal vector estimation phase, and trained otherwise similar models on the images.

As the results in Table~\ref{tab:ablation_NN} show, the nearest-neighbour parameter does not affect NormalView's performance greatly. On dense MLS~data, the best results are achieved with \(NN = 3\), with \(NN = 20\) being close behind. On MLS~data, the results are all close to each other, with the exception of \(NN=30\), where performance is noticeably weaker. On the ALS~data the results are even more homogenous, with \(NN = 5\) being the best, followed again closely by \(NN=20\). However, the metrics of the various models are extremely similar. The results indicate that NormalView is robust to the choice of the nearest-neighbor parameter in normal vector estimation. This is important, as it implies that extensive parameter optimisation is not necessary when applying the method.

We studied the effect of point density on the performance of NormalView on the MLS dataset. We conducted the study by randomly subsampling the MLS tree segments to a desired point density, and creating images from the subsampled segments, and training models on them.

The results, presented in Table~\ref{tab:point_density}, are perhaps a little surprising. The performance degrades as the point density decreases, but only marginally. Even on the lowest density data, which has a point density of \qty{500}{pts/m^2}, the overall accuracy drops only 2.4 percentage points, while the MAA drops by 4.9 percentage points. On \qty{2000}{pts/m^2} the OA drops only by 0.6 percentage points and the MAA and \mafone \ decrease by 3.1 and 1.0 percentage points, respectively, when compared to the model trained on the full density data. The separation between the models trained on subsampled data is even more moderate. The results on the downsampled data for point densities \qty{2000}{pts/m^2} and \qty{1000}{pts/m^2} are slightly stronger than what NormalView achieved on the ALS data and seven species, which is a dataset consisting of the same species at approximately the same point density. The above findings suggest that a full coverage of the tree, as obtained from a below-canopy scanner, is more important for performance than a high point density. However, from the MAA and \mafone \ it can be inferred that it is easier to distinguish minority species when the point density is higher.

Finally, Table~\ref{tab:no_slice} shows the metrics on models trained on intensity, normal, and black-and-white MLS~data without slice images. The slice images clearly benefit the intensity and NormalView models, particularly the latter, where removing slice images decreases the macro-average accuracy by 3.7 percentage points. This indicates that exposing the internal branch and trunk structure provides additional discriminative geometric information beyond the outer silhouette. In contrast, the black-and-white model slightly benefits from removing slice images, suggesting that silhouette-only representations cannot effectively exploit the additional internal structure and may instead be affected by the increased image complexity.

\subsection{Effect of image size on model performance}\label{sub:imsize}

\begin{figure}[!ht]
\captionsetup[subfigure]{skip=1pt, singlelinecheck=false}

    \centering
    \begin{subfigure}[b]{0.495\textwidth}
        \centering
        \caption{Metrics of MLS models trained with varying image size.}
        \includegraphics[width=\textwidth]{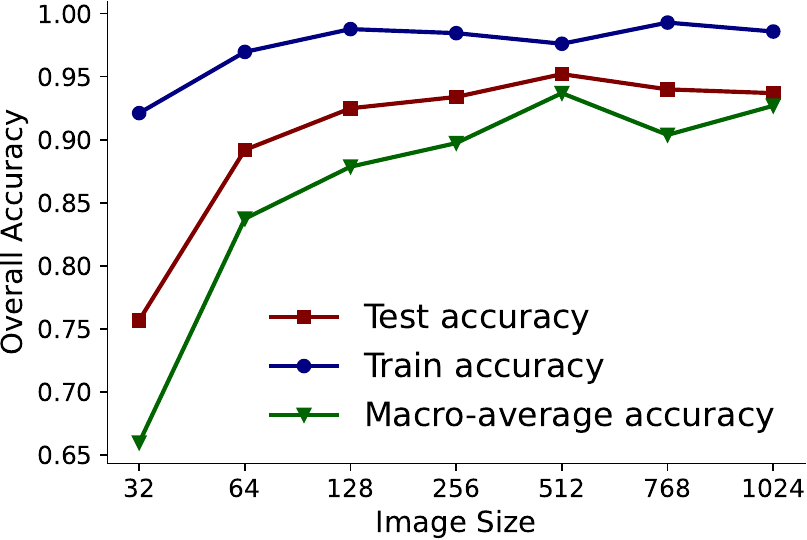}
        \label{fig:imsz}
    \end{subfigure}
    \hfill
    \begin{subfigure}[b]{0.495\textwidth}
        \centering
        \caption{Proportion of empty pixels in images created from MLS data.}
        \includegraphics[width=\textwidth]{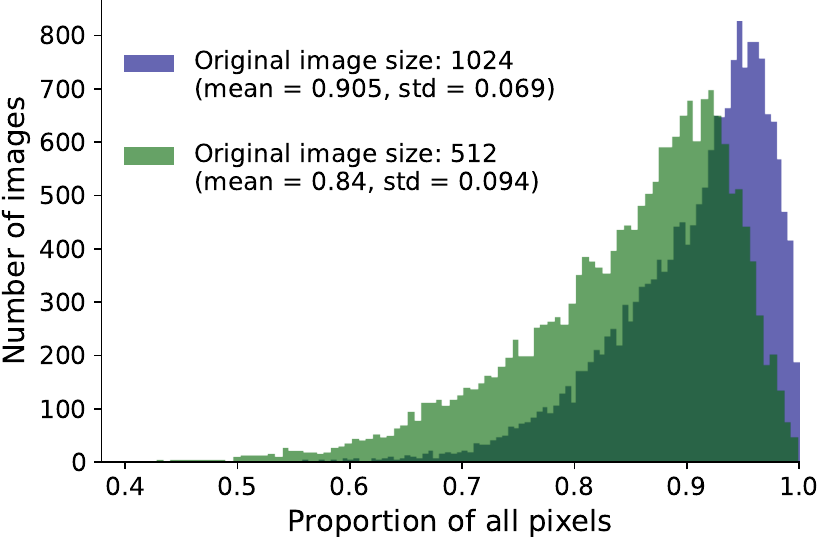}
        \label{fig:emptypixel}
    \end{subfigure}
    \caption{(a) Performance metrics of models trained on MLS data and varying image sizes. (b) Proportions of empty pixels in images created from MLS data to sizes \(1024\times1024\) and \(512\times512\).}
    \label{fig:imsizefigs}
\end{figure}

Figure~\ref{fig:imsz} shows that increasing image size improves model performance to a certain extent. The highest test accuracy (95.2\,\%) and macro-average test accuracy (93.8\,\%) are achieved by the model using 512 as the image size. The models using higher image size, 768 and 1024, perform slightly worse than the model using size 512, and rather equally to the model using size 256. The training accuracies are markedly higher than test accuracies for all models. Curiously, the model using size 512 has the third-worst training accuracy, surpassing only the models using sizes 32 and 64. This hints that the model overfitted to the training data the least among the models. The model using image size 512 achieves better results than the model trained on intensity-based images, where the initial image size is set to 512, which are visible in Table~\ref{tab:MLS_mod_res}. Computationally, the models using larger image sizes require more memory and processing power, both during training and inference. 

Figure~\ref{fig:emptypixel} shows that the images created into size \(1024\times1024\) tend to have more empty pixels. The pixel sizes are \qty{1.64}{cm} (\(\pm \ \qty{0.47}{cm}\)), and \qty{3.28}{cm} (\(\pm \ \qty{0.93}{cm}\)) for the sizes 1024 and 512, respectively, when evaluated by the average heights of the trees.

\subsection{Further discussion}\label{sub:furtheranalysis}

The results in section ~\ref{sub:Model_perf}, as well as Figure~\ref{fig:mlsconf}, indicate that NormalView is highly adept at classifying tree species from high-density MLS data. Especially the confusion matrices~\ref{fig:mlsconf} show that NormalView takes minority species, rowan and lime, into account better than intensity-based models, or models using black-and-white images. Compared to the model trained on black-and-white images, NormalView achieved a $4.1$ percentage points better MAA and $3.3$ percentage points better \mafone, see Table~\ref{tab:MLS_mod_res}. This implies that if we only have geometric data on hand, the normal vector colouring improves classification performance. In particular, it seems that the normal vector colouring captures local information, which yields better classification performance compared to the black-and-white model, which uses only the silhouette information.

\begin{figure*}[ht]
    \centering
    \includegraphics[width=\textwidth]{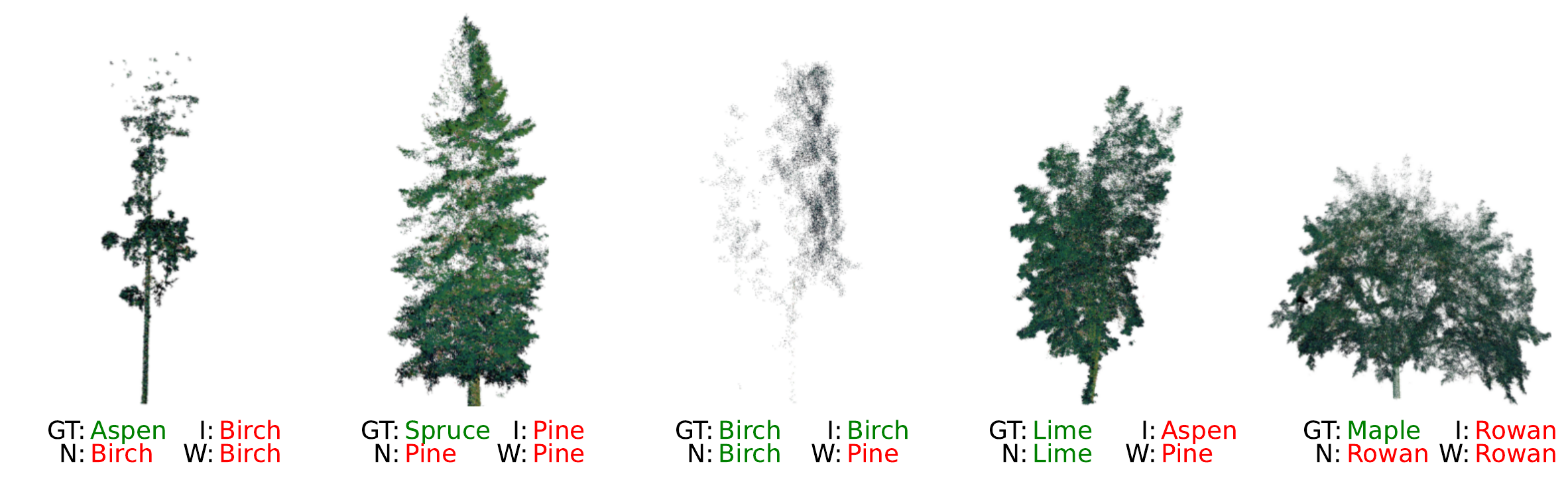}
    \caption{Examples of MLS-trees misclassified by at least one model. GT refers to ground truth, I, N and W refer to the predictions given by the intensity, normal and black-and-white models respectively. Green indicates a correct prediction and red an incorrect one.}
    \label{fig:misclassification}
\end{figure*}

On the full ALS dataset, models using multiple intensity channels proved to be the best. The model with all three intensity channels was a strong performer with the seven species and the nine species datasets. This is in accordance with the findings in~\citet{taher2025}, where they found that especially low-density point cloud data benefits from multispectral information. On lower-density data, the importance of radiometric information becomes more significant, as the data is sparser, and the underlying geometry of the point cloud becomes inconsistent and more unreliable. The performance of the best models on the data is comparable to the best models of the benchmark conducted by~\citet{taher2025}, where they achieved an OA (MAA) of \(92.0\,\%\) (\(85.1\,\%\)) on a similar train-test split as here. Our best-performing model, the one with all three intensity channels, shows essentially equal performance. The corresponding metrics for NormalView are \(91.8\,\%\) (\(79.1\,\%\)), which are worse, but not by a large margin. The key difference is that the best model in the benchmark utilised full radiometric data from all three scanners, while NormalView uses only geometric information. Images coloured by the normal vector estimations benefit from the tree point clouds having consistent and high-density trunks and branches, as this allows for smoothly  varying and consistent normal vectors and hence smoother and bounded colour gradients. This likely explains the comparatively stronger performance of NormalView on MLS data, compared to ALS data. On the ALS dataset with the same seven species as in the MLS data, NormalView was not the strongest performer, unlike on the MLS data. The results on the MLS data suggest that especially in applications requiring high-density point clouds, such as exact inventory of natural forests, normal vector colouring provides benefits.

There seems to be little performance difference between the models using ALS data and intensity information from a single channel. On the seven species, the model using channel 1 (VUX-1HA) was the best performer amongst the single channel models, while on nine species the model using channel 2 (miniVUX-1DL) achieved the strongest results. Overall, the performance results are similar and there seems to be little difference in which scanner is used. This suggests that using radiometric information with projection-based methods can be made viable on scanners with varying properties, such as the wavelength of the laser pulse.

We see from Figure~\ref{fig:ALS_species_metrics} that across the board the models perform the worst, by precision and \(\mathrm{F}_1\), on oak, and, to a lesser degree, alder. These species are the two with the smallest representation in our dataset, and hence the least variety.  In addition, the segments of the alders and oaks have some impurities, as suggested
by Figure 5 in~\citet{taher2025}. Alders tend to grow in groups, making segmentation more
difficult. A significant portion of oak segments are only partial segments of the trees. The low
performance on these species can probably at least partially be attributed to low sample counts
and lower-quality segments. From the confusion matrices in~\ref{app:confmat} we see that almost all models tend to systematically classify aspens as birches. Birches and aspens have rather similar silhouettes, which explains the errors. Similarly, rowans seem to get classified as maples and birches. Some examples of trees misclassified by the MLS models are shown in Figure~\ref{fig:misclassification}. It seems that poorly captured trees with lacklustre silhouettes are prone to misclassification, such as the aspen and birch trees in Figure~\ref{fig:misclassification}. This is not the only explanation for misclassification, as for example the spruce, lime, and maple trees in Figure~\ref{fig:misclassification} are captured well by the scanner, and have full silhouettes. To enhance the performance of the models on species with similar silhouettes, one could, for example, include more detailed bark information, as is done in the DetailView method discussed by~\citet{Puliti}.

The $\mathrm{F}_1$-scores for aspens are satisfactory across all data modalities and models. On the MLS data, the scores are around 80\,\%, while between 83 and 87\,\% for the ALS data and all nine species. This shows that we are able to map aspens with high confidence from both MLS and ALS data. The use of multispectral information improves the scores by a small margin, but the intensity-free NormalView and black-and-white models achieve \(\mathrm{F}_1\) scores of 85\,\% and 83\,\% respectively, highlighting the capabilities of projection-based methods.

The black-and-white model is the worst performing model on MLS data, and among the worst performers on ALS data. On the MLS data and the nine species ALS data, it achieves an OA (MAA) of $94.0\,\%$ ($90.7\,\%$) and $91.6\,\%$ ($79.5\,\%$) respectively. These results show that even using purely black-and-white orthographic projections allows us to classify tree species on a reasonably high level. This is further supported by the performance of the black-and-white model on the MLS + ALS data. The performance is due to the high capabilities in which modern image-based architectures, such as YOLO, are able to extract meaningful representations and features from the data. The image classification backbones used in previous studies of tree species classification seem to be, in general, slightly outdated, such as the VGG-11~\citep{VGG-11} used by~\citet{Marinelli}, or the DenseNet-201~\citep{densenet} used by the DetailView method in~\citet{Puliti}.

An important finding of this study is that the relative advantages of the different projection methods depend on the acquisition modality. On dense MLS data, NormalView consistently outperforms the intensity-based and silhouette-based approaches, demonstrating the value of detailed local geometric information. On ALS data, however, multispectral intensity information proves more beneficial, particularly for distinguishing minority species in the nine-species dataset. When both acquisition methods are combined into a single dataset, the performance differences between the models largely disappears. This suggests that modality-specific variation limits the effectiveness of representations tailored to a single sensor type and highlights the importance of developing acquisition-invariant features for cross-platform tree species classification.

The ablation studies provide additional insight into the robustness of the proposed approach. NormalView is found to be largely insensitive to the choice of the nearest-neighbour parameter used for normal estimation, suggesting that the method does not require extensive parameter tuning when applied to new datasets. Similarly, the relatively small performance degradation observed after substantial point cloud downsampling indicates that the method primarily exploits larger-scale structural characteristics of trees rather than relying on extremely dense point clouds. This is encouraging from a practical perspective, as high-density terrestrial scans are not always available. Furthermore, the experiments demonstrate that slice images provide meaningful additional information for the intensity-based and geometry-based representations by exposing the internal branch and trunk structure, whereas silhouette-only images benefit less from the additional views. Together, these findings indicate that NormalView is not only accurate but also robust with respect to several practical factors encountered in real-world applications.

As seen in Section~\ref{sub:imsize}, using an appropriate image size can significantly improve the predictive power of the networks. Using larger images allows for more detailed information, at the cost of longer training times and higher hardware requirements. Hence, it is critical that before training the models, we have some notion of the appropriate size into which to create the images. For images created into size \(1024\times1024\), one pixel size corresponds approximately to a square of size \(\qty{1.64}{cm} \times \qty{1.64}{cm}\) in the projective plane. In the preprocessing phase, we subsampled the point clouds to a spatial resolution of~\qty{2}{cm}, which is larger than the size of the pixels. As a result, there are many empty pixels, even at points where we would assume the tree to be visible. Certainly, as a general guideline, the resolution should be such that one pixel encompasses an area that is slightly larger than the spatial resolution of the underlying point cloud data. This is supported by our findings in Section~\ref{sub:imsize}, where the model with image size 512 was the best. For images of size \(512\times512\), the pixel size is \(\qty{3.28}{cm} \times \qty{3.28}{cm}\), which is roughly one-and-a-half times larger than the spatial resolution of the densest parts of the point clouds. Figure~\ref{fig:emptypixel} shows that higher resolution images tend to have more empty pixels, which is, of course, expected. Convolutional neural networks (CNNs) are usually designed with the assumption that the input data will be dense~\citep{graham2018}. If the image size in the projections is large, the images will start to be rather sparse, in the sense that most of the image will be empty pixels, containing no information. At the very least, this will waste time and require more computational resources with little to no added performance benefit. In the worst case, empty pixels will negatively affect the model. 

\subsection{Limitations}\label{sub:limits}

The datasets used in this study, while containing the most predominant boreal tree species, are incomplete even for traditional Finnish ecosystems. Hence it is not applicable to use the models directly in practice. The training of the models also requires significant amounts of annotated data. Collecting and labelling training data for deep learning models is notoriously time-consuming and, as such, highlights the need for large multispecies datasets, such as the recently released FORspecies20k and the MS-ALS-SPECIES dataset~\citep{forspecies20k, espooDATA}.

In the MLS data, the segment quality is extremely high. Our empirical experience is that better segment quality improves classification performance. For real world applications, the segment quality might be significantly lower, and the accuracy of the models may suffer from these inaccuracies. The effect of segment quality on tree species classification warrants further study.

Due to the computational cost of training the models, each experiment was conducted using a single train-test split and a single training run. Future work could investigate the variability of the results through repeated runs with different random seeds or cross-validation. 

\section{Conclusions}\label{sec:conclusions}

In this study, we presented NormalView, a projection based tree species classification method designed for robust classification from any MLS or high-density ALS data---even in the absense of intensity information. The method embeds local geometric information of tree point clouds into projection images via colour. The colour values are derived from normal vector estimates of the points.

We compared our projection-based method with models trained on projection images using intensity information and models using black-and-white projection images. We found that on high-density seven species MLS data our NormalView achieves superior performance compared to the other methods, with the overall accuracy (macro-average accuracy) reaching 95.5\,\% (94.8\,\%). The strong performance on dense MLS data, together with the robustness observed in the ablation studies, suggests that NormalView is particularly well suited for forestry related point cloud processing pipelines with high-density data. 

For high-density helicopter ALS data, we found that our method achieves comparable performance with methods using intensity information. However, the best model on the ALS data was the one using intensity information from three channels of the multispectral ALS. It achieved an overall (macro-average) accuracy of 92.5\,\% (84.9\,\%) on the full dataset with nine species. These results are on par with the best models in~\citet{taher2025}, showing that projection-based methods using strong CNN-backbones can achieve performance comparable to that of state-of-the-art 3D models. The experiments indicate that geometry-based representations are most advantageous on dense point clouds, while multispectral radiometric information remains beneficial for more challenging low-density airborne datasets.

Our tests show that using multispectral data, i.e. combining radiometric information from multiple scanners, improves the prediction ability of projection-based models. The benefits are best seen for species belonging to minority classes. We also inspected the effect of image size on model performance, and found that using larger images is beneficial up to a certain point, enhancing the accuracy of the model. For the high-density MLS data we found that image size \(512\times 512\) produces the best results.

\section*{Data availability}
The labeled MLS point clouds and model weights are available in \citep{MLSspeciesDATA}, under Creative Commons Attribution-NonCommercial 4.0 International licence. Commercial rights can be granted separately for collaborating companies by FGI.

\section*{Code availability}
The code used for NormalView image creation and model training is available in \citep{NormalView_github}.

\section*{CRediT authorship contribution statement}

\textbf{Juho Korkeala:} Writing - Original Draft, Conceptualization, Methodology, Software, Investigation, Data Curation, Visualisation. \textbf{Jesse Muhojoki:} Writing – Original Draft, Resources, Supervision. \textbf{Josef Taher:} Writing – Review \& Editing, Conceptualization. \textbf{Klaara Salolahti:} Writing – Review \& Editing, Data Curation, Visualisation. \textbf{Matti Hyyppä:} Writing – Review \& Editing, Data Curation. \textbf{Antero Kukko:} Writing – Review \& Editing, Project Administration, Funding Acquisition, Supervision. \textbf{Juha Hyyppä:} Writing – Original Draft, Project Administration, Funding Acquisition, Supervision.

\section*{Acknowledgments}

The study has received funding from the Research Council of Finland (RCF), and the NextGenerationEU instrument through the following grants "Collecting accurate individual tree information for harvester operation decision making" (RCF 359554), "Digital technologies, risk management solutions and tools for mitigating forest disturbances" (RCF 353264, NextGenerationEU), and "Forest-Human-Machine Interplay -- Building Resilience, Redefining Value Networks and Enabling Meaningful Experiences" (RCF 359175) using RCF research Infrastructure “Measuring Spatiotemporal Changes in Forest Ecosystem” (346382)".

\section*{Declaration of competing interest}
The authors declare that they have no known competing financial interests or personal relationships that could have appeared to influence the work reported in this paper.

\scriptsize{\bibliography{references}}

\normalsize
\begin{appendices}
\counterwithin{figure}{section}
\counterwithin{table}{section}
\onecolumn

\section{Confusion matrices}\label{app:confmat}

\begin{figure*}[ht]
    \centering
    \includegraphics[width=\textwidth]{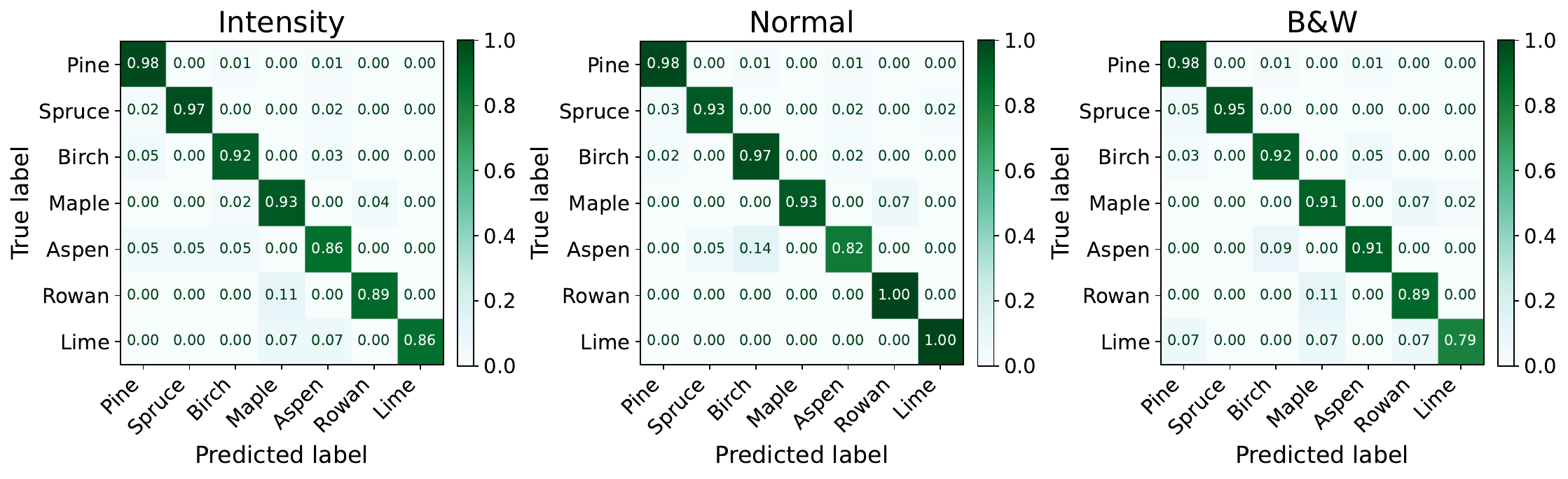}
    \caption{Normalised MLS confusion matrices}
    \label{fig:mlsconf}
\end{figure*}

\begin{figure*}[ht]
    \centering
    \includegraphics[width=\textwidth]{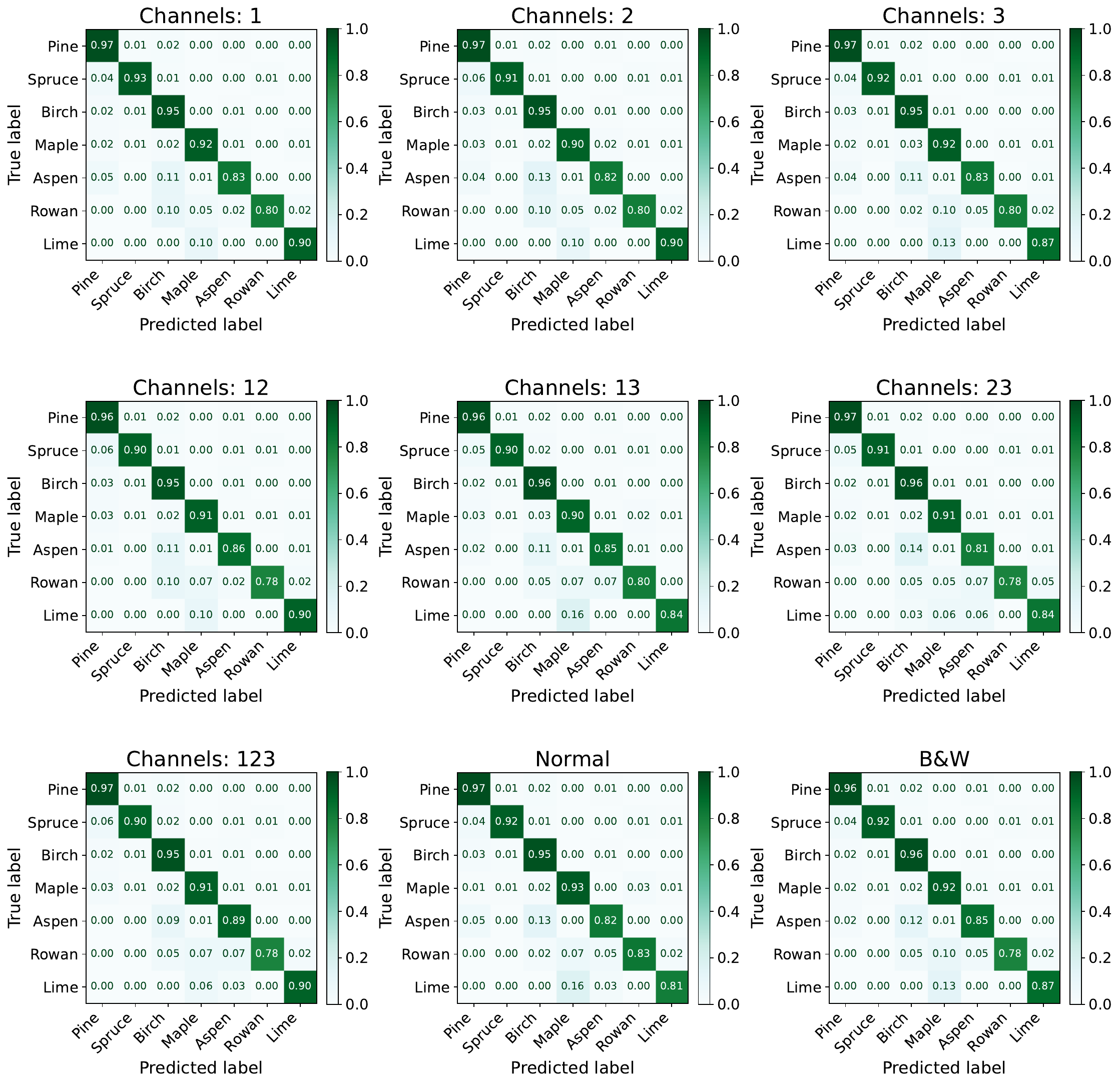}
    \caption{Normalised confusion matrices (ALS, 7 species)}
    \label{fig:alsconf}
\end{figure*}

\begin{figure*}[ht]
    \centering
    \includegraphics[width=\textwidth]{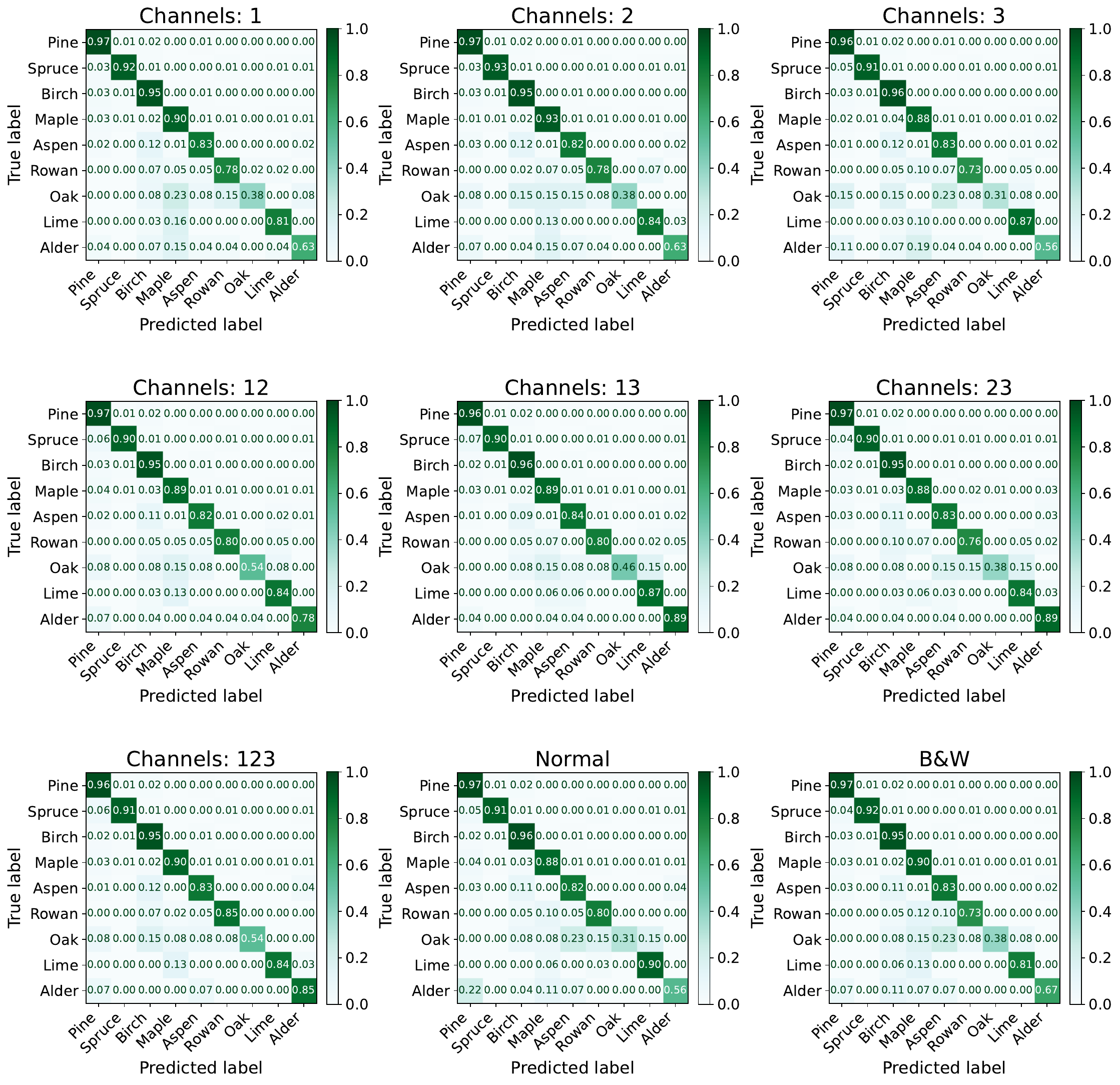}
    \caption{Normalised confusion matrices (ALS, 9 species)}
    \label{fig:als9conf}
\end{figure*}

Figures~\labelcref{fig:mlsconf,fig:alsconf,fig:als9conf} present the normalised confusion matrices for the trained models.

\begin{figure}[ht]
    \centering
    \includegraphics[width=\linewidth]{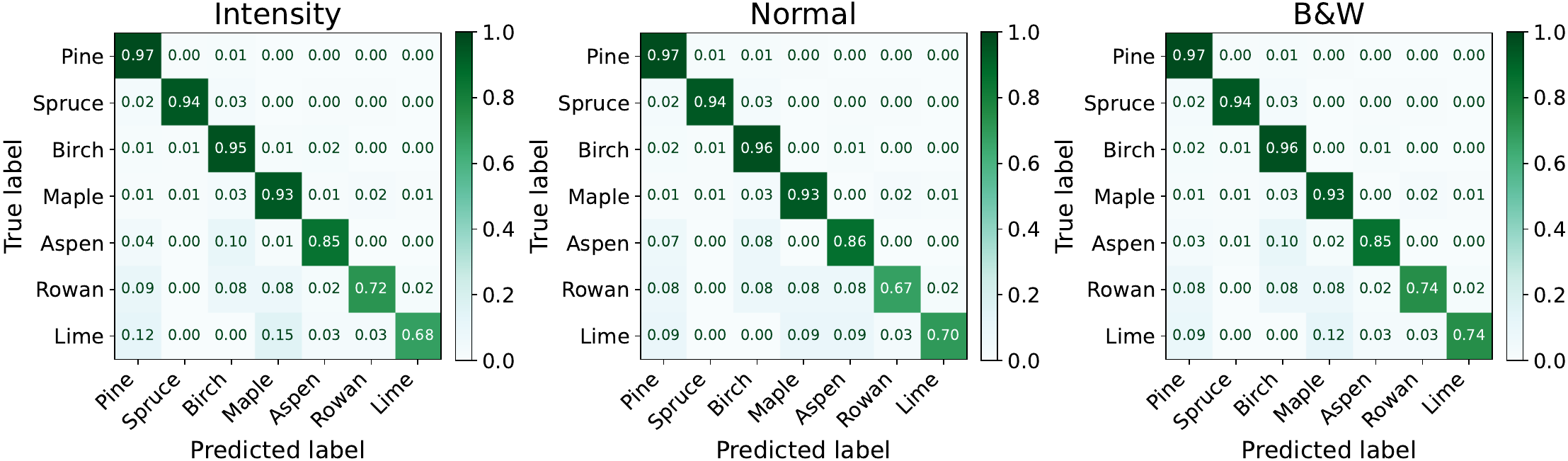}
    \caption{Normalised confusion matrices (MLS + ALS, 9 species)}
    \label{fig:MLSALSconf}
\end{figure}

\end{appendices}

\end{document}